%% file: main_v2.tex
\documentclass{article}

\usepackage[preprint]{corl_2026} 

\usepackage[numbers]{natbib}

\input{math_commands.tex}

\usepackage{subfigure}
\usepackage{amsmath}
\usepackage{amssymb}
\usepackage{mathtools}
\usepackage{amsthm}
\usepackage{array}
\usepackage{algorithm}
\usepackage{algpseudocode}
\usepackage{graphicx}
\usepackage{subcaption}
\usepackage{pdflscape}
\usepackage{pdfpages}
\usepackage{wrapfig}
\usepackage{makecell}
\usepackage{mine}
\usepackage[para]{footmisc}

\title{Morphology-Conditioned World Model for Cross-Embodiment Quadrupedal Locomotion}

\author{
\textbf{Mohamad H. Danesh}$^{1,2}$ \quad 
\textbf{Chenhao Li}$^{3}$ \quad
\textbf{Amin Abyaneh}$^{1,2}$ \quad
\textbf{Anas Houssaini}$^{1,2}$ \\
\textbf{Kirsty Ellis}$^{2,4}$ \quad
\textbf{Glen Berseth}$^{2,4}$ \quad
\textbf{Marco Hutter}$^{3}$ \quad
\textbf{Hsiu-Chin Lin}$^{1,2}$\\\\
$^{1}$McGill University, Canada \quad
$^{2}$Mila - Quebec AI Institute, Canada \\
$^{3}$ETH Zurich, Switzerland \quad
$^{4}$Universite de Montreal, Canada
}

\begin{document}
\maketitle

\begin{abstract}
World models promise a paradigm shift in robotics, where an agent learns the physics of its environment once and then acquires behaviors efficiently. Yet the learned dynamics models at their core are typically morphology locked. In legged locomotion, a dynamics model trained on an ANYmal-D quadruped fails on a Unitree Go1 because it overfits to one robot's embodiment rather than capturing the locomotion dynamics shared across robots, so even a small change in actuator dynamics or limb length forces retraining from scratch. However, if we formalize a robot's unique physical traits into a morphology specification, a controller for a family of robots can utilize this blueprint in two ways. It can feed the specification to a model-free policy, or it can feed the specification to a learned dynamics model and extract the policy in imagination. We argue for the second route and introduce the Quadrupedal World Model (\ourMethod), which conditions a single generative dynamics model on scale-invariant physical features and trains policies entirely inside it, through a physical morphology encoder, an adaptive reward normalizer, and morphology conditioning in the latent dynamics. Holding the morphology information identical, a model-free policy matches \ourMethod on the training cohort but degrades on unseen morphologies, while \ourMethod transfers zero-shot with no fine-tuning, adaptation, or warm-up in such cases. To our knowledge, this is the first world model to demonstrate zero-shot cross-embodiment transfer within the quadrupedal family.
\end{abstract}

\section{Introduction}\label{sec:intro}

Robot learning is data hungry, and data is mostly collected on a single hardware platform, so the resulting controllers and the dynamics models on which they rely overfit to specific kinematic and dynamic properties \citep{liu2025compass}. Upgrading motors or deploying across a heterogeneous fleet then means collecting new samples and retraining from scratch \citep{Yarats_Zhang_Kostrikov_Amos_Pineau_Fergus_2021, Kostrikov-RSS-23}. Large scale simulations \citep{NVIDIA_Isaac_Sim, mittal2025isaaclab, Genesis} ease the wall-clock cost of training but do not remove the coupling between a learned model and its training hardware. Furthermore, blind domain randomization without informative conditioning tends toward conservative control rather than performance tuned to the robot at hand \citep{danesh2025safe}. Our starting point is a simple insight. A useful model of locomotion should capture not how \textit{one} robot moves, but how robots within \textit{a morphological family} move. This shared understanding allows a single model to leverage training data across different scales, ensuring that a hardware change no longer requires relearning dynamics from scratch.

Such a model must know which robot it is controlling, since across a family the transition dynamics vary mostly based on the embodiment. We summarize a robot by a morphology vector $\mu$ that encodes its kinematic constraints, mass distribution, and actuator limits. We take $\mu$ to be given by the robot's engineering description, since a nominal specification such as a Universal Scene Description already records the geometry and inertial properties that distinguish one platform from another. When such a description is unavailable, $\mu$ is instead inferred from interaction history, often to refine an imperfect prior \citep{kumar2021rma, yu2017preparing}, a setting that is complementary to ours rather than competing with it. We study the regime in which an accurate $\mu$ is available, where the design question we care about is cleanest.
The question is then not how to obtain $\mu$ but how it should be used. One option feeds $\mu$ to a model-free policy and learns the controller directly \citep{chen2018hardware, xiong2023universal}. The other feeds $\mu$ to a dynamics model and recovers the controller by training in imagination on that model \citep{sutton1991dyna, wu2022daydreamer}.

\begin{figure}[t]
  \centering
  \includegraphics[width=\textwidth]{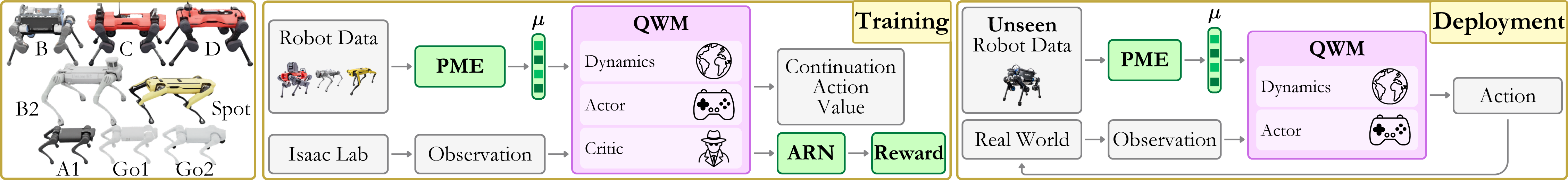}
    \caption{Overview of \ourMethod. \textbf{Left}: The heterogeneous quadrupedal morphology cohort. \textbf{Middle}: Training pipeline, where robot data is processed by PME (\autoref{sec:morph_embodiment}) to extract the morphology vector $\mu$, which together with environment observations trains \ourMethod, while ARN (\autoref{sec:hetero_rew}) keeps the reward signals consistent across diverse platforms. \textbf{Right}: Deployment pipeline, where for an unseen robot the same PME module from training extracts its morphology vector $\mu$.}
\label{fig:overall_quads}
\end{figure}

We argue that for cross-embodiment generalization, the second option is the better utilization of $\mu$. The dynamics of a legged robot are governed by physics and vary smoothly with morphology, since changing a limb's length or mass moves the equations of motion continuously, whereas the optimal control law for those dynamics need not vary smoothly, since small physical changes can reshape a gait. A model that learns the shared dynamics of a family and is conditioned on $\mu$ can therefore synthesize a coherent simulator for an unseen morphology by interpolating in physical feature space, and a policy re-derived against that simulator inherits the generalization, while a policy that maps $\mu$ directly to actions must generalize a rougher function. The obstacle is that existing world models (WMs) are morphology locked, a model trained on a heavy robot overfits to its dynamics and fails on a lighter one \citep{hafner2025mastering, silver2017mastering, hafner2019learning, hafner2022mastering, wu2022daydreamer, li2025robotic}. Our contribution is to make the model-based route work across a family, capturing the shared physics of locomotion while conditioning on what differs between robots, following the roadmap for \textit{Type III Robotic Agents} \citep{fung2025embodied}.

We take a first step toward a generalizable Quadrupedal World Model (\ourMethod), a morphology-conditioned dynamics model for a quadrupedal kinematic family (\autoref{fig:overall_quads}). \ourMethod comprises three components: a fusion architecture that injects morphology features into the WM, a Physical Morphology Encoder (PME) that derives scale-invariant features $\mu$ from robot descriptions, and an Adaptive Reward Normalizer (ARN) that balances learning signals across robots with heterogeneous reward scales. To isolate the value of the model-based route, our comparison holds morphology information fixed and contrasts \ourMethod against model-free baselines conditioned on identical $\mu$. The two match on the training cohort, yet across unseen morphologies, parameter inaccuracy, and unmodeled dynamics, \ourMethod generalizes significantly better, including zero-shot transfer to hardware.

\section{Related Work}\label{sec:related}
\textbf{MBRL} improves sample efficiency by learning predictive environment models \citep{sutton1991dyna, moerlandnow}. While early methods relied on Gaussian processes \citep{deisenrothpilco}, modern approaches scale to complex environments by planning in latent spaces \citep{ha2018world}. Notably, the Dreamer family \citep{hafner2019learning, hafner2022mastering, hafner2025mastering} demonstrated that policies can be learned entirely in imagination, with DreamerV3 \citep{hafner2025mastering} providing robustness via symlog predictions and KL balancing. This shift toward WMs is supported by theory suggesting they are inherent to general agents \citep{richens2025general}. As the paradigm matures, the foundational RSSM architecture \citep{hafner2019learning} is being replaced by Transformers \citep{chen2024transdreamer, dedieuimproving}, discrete autoencoders \citep{micheli2023transformers, micheli2024efficient}, stochastic transformers \citep{zhangstorm}, contrastive coding \citep{burchi2025learning}, and linear attention or state-space models \citep{wang2025drama, wang2024parallelizing}. Within robotics, MBRL has enabled significant milestones, starting with DayDreamer's \citep{wu2022daydreamer} online learning and expanding into long-horizon prediction \citep{li2025robotic}, visual reconstruction \citep{lai2024world, zhangdymodreamer}, and offline-to-online adaptation via epistemic uncertainty \citep{li2025offline}. Today, applications span sim-to-real distillation of privileged WMs \citep{yamada2023twist}, humanoid terrain traversal \citep{zheng2025huwo}, manipulation \citep{ferraro2023focus, bi2023sampleefficient}, safety \citep{huang2024safedreamer}, and robust exploration \citep{sekar2020planning, qiao2024bounded, zollicoffernovelty}. Yet, despite these varied successes, existing models remain fundamentally tied to a single embodiment. Overcoming this limitation, by learning dynamics that transfer across morphological changes, is the precise gap our work targets.

\textbf{Generalist and cross-embodiment control} scales a single controller across tasks or robots. TD-MPC2 \citep{hansen2024tdmpc2} learns latent task embeddings, PWM \citep{georgiev2025pwm} extracts policies via first-order optimization through differentiable WMs, and others adapt from offline data \citep{zhao2025efficient, zhang2025prelar, feng2023finetuning}. In locomotion, several methods rely on analytical priors or heuristic skill selection \citep{shafiee2023manyquadrupeds, feng2022genloco, wu2025vocaloco}. In contrast, \ourMethod operates directly in joint space by mapping robot descriptions to dynamics, enabling planning without analytical inverse kinematics. Furthermore, where prior multi-robot methods must balance heterogeneous objectives via complex loss weighting \citep{ma2024harmonydream, leecqm}, we bypass this complexity using a lightweight per-robot reward Exponential Moving Average (EMA).

\textbf{Morphology representation} for cross-embodiment generalization largely relies on architectural biases. GNNs model the kinematic tree \citep{wang2018nervenet, whitman2021learning, huang2020one}, sometimes with symmetry constraints \citep{xie2025morphologicalsymmetryequivariant, wei2025msppo} or heterogeneous nodes \citep{butterfield2025mihgnn}, while transformers tokenize the kinematic chain \citep{gupta2022metamorph, bohlinger2025one, xi2025unilegs, trabucco2022anymorph, sferrazza2024body}, scaling to thousands of procedural embodiments \citep{ai2025embodiment}. Others infer morphology implicitly from interaction history \citep{kumar2021rma, yu2017preparing, pathak2019learning} or learn latent action spaces \citep{wang2024crossembodiment, jang2025dreamgen, li2025robotic, zhang2025crossembodiment}, which suits settings without an explicit description but requires a warm-up interaction phase before deployment, the generalization gap \citet{fung2025embodied} flag as critical. Since we target the regime where an accurate description is available, we instead use explicit parametric encoding \citep{chen2018hardware, mishra2025mcarl, xiong2023universal}, computing scale-invariant features fed directly into the WM for zero-shot adaptation without the overhead of GNNs.

\section{Background}\label{sec:background}
We formulate locomotion as a Partially Observable Markov Decision Process $(\mathcal{S}, \mathcal{O}, \mathcal{A}, \mathcal{P}, \mathcal{R}, \gamma)$, where $\mathcal{S}$ denotes the simulator state space, $\mathcal{O}$ the observation space, $\mathcal{A}$ the action space specifying target joint positions, $\mathcal{R}(s_t, a_t)$ the reward function producing reward $r_t$, and $\gamma \in [0,1)$ the discount factor. The transition dynamics are defined by a stochastic transition function $\mathcal{P}(s_{t+1} \mid s_t, a_t)$, which specifies the probability distribution over next states given the current state and action. At time $t$, the agent receives a partial observation $o_t \in \mathcal{O}$ consisting of joint positions, joint velocities, base linear velocity, angular velocity, projected gravity, velocity commands, and previous actions.

\subsection{Recurrent State-Space Models}
Our WM backbone is based on the Recurrent State-Space Models (RSSM) introduced in the Dreamer family of algorithms \citep{hafner2025mastering, hafner2019learning, hafner2022mastering}.
RSSMs enable learning in partially observable environments by decomposing the latent state into a deterministic recurrent state $h_t$ that aggregates history (short-term memory) and a stochastic state $z_t$ that captures information about the current observation (spatial representation). Parameterized by $\phi$, these components evolve through a deterministic function $f_\phi$ and the probability distributions $p_\phi$ and $q_\phi$:
{\subfootnotesize
\begin{align}
    \text{Recurrent Model:} & \quad h_t = f_\phi(h_{t-1}, z_{t-1}, a_{t-1}) \label{eq:rnn} \\
    \text{Representation Model:} & \quad z_t \sim q_\phi(z_t \mid h_t, o_t) \label{eq:post} \\
    \text{Transition Predictor:} & \quad \hat{z}_t \sim p_\phi(\hat{z}_t \mid h_t) \label{eq:prior} \\
    \text{Decoder, Reward \& Continuation:} & \quad \hat{o}_t, \hat{r}_t, \hat{c}_t \sim p_\phi(\cdot \mid h_t, z_t) \label{eq:rew}
\end{align}}
The dynamics evolve through the \textit{Recurrent Model} (\autoref{eq:rnn}), which updates the memory $h_t$ based on the previous action. During training, the \textit{Representation Model} (\autoref{eq:post}) infers the posterior state $z_t$ using the observation $o_t$, whereas during planning, the \textit{Transition Predictor} (\autoref{eq:prior}) predicts the prior $\hat{z}_t$ solely from history. To ground these latents, the model reconstructs observations via the \textit{Decoder} and predicts training signals via the \textit{Reward} and \textit{Continuation} heads (\autoref{eq:rew}).

The model parameters $\phi$ are optimized by minimizing a variational bound that includes reconstruction of observations $o_t$, prediction of the reward $r_t$ and continuation signal $c_t \in \{0, 1\}$, and a KL-divergence regularizer that forces the prior (imagination) to align with the posterior (reality):
\begin{equation}
\resizebox{0.9\linewidth}{!}{$
\mathcal{L}_{\text{WM}} = \mathbb{E}_{q} \Big[
\underbrace{
-\ln p_\phi(o_t \mid h_t, z_t)
- \ln p_\phi(r_t \mid h_t, z_t)
- \ln p_\phi(c_t \mid h_t, z_t)
}_{\text{Reconstruction, Reward \& Continuation}} + \beta
\underbrace{
\text{KL}\!\left[q_\phi(z_t \mid h_t, o_t)\| p_\phi(\hat{z}_t \mid h_t)\right]
}_{\text{Dynamics Consistency}}
\Big]
$}
\end{equation}

\subsection{Learning Policy in Imagination}
A key advantage of WMs is the decoupling of policy learning from physical interaction. The policy $\pi_\theta(a_t | h_t, z_t)$ and critic $V_\psi(h_t, z_t)$, parameterized by $\theta$ and $\psi$ respectively, are trained entirely on \textit{imagined trajectories} rolled out by the transition predictor.
Starting from an initial state, the model interacts with itself by sampling $a_t \sim \pi_\theta(\cdot \mid h_t, \hat{z}_t)$, where $h_{t+1}$ and $\hat{z}_{t+1}$ come from \autoref{eq:rnn} and \autoref{eq:prior}, respectively.
Because the policy is trained only against the model's latent dynamics, its behavior on a new robot is as good as the model is for that robot. This shifts the burden of cross-morphology transfer from the policy onto the dynamics model, and it is the property we exploit. When the model represents the new robot's dynamics properly, the policy can be rolled out zero-shot in imagination and deployed without further interaction.

\section{Quadrupedal World Model}\label{sec:qwm}
We build upon DreamerV3 \citep{hafner2025mastering}, modifying the encoder and transition dynamics to be conditioned on $\mu$.
For a family of robots, the transition is no longer fixed but varies with the robot, which we write as $\mathcal{P}_\mu(s_{t+1} \mid s_t, a_t)$ for a morphology variable $\mu \in \mathcal{M}$ that ranges over quadrupedal embodiments.
\ourMethod adds three components. The first is a morphology-conditioned WM that fuses static morphology features with dynamic proprioception. The second is the PME, which derives those features from robot description files. The third is an ARN scheme that stabilizes learning across heterogeneous robots, given their various reward scales.

\subsection{Morphology-Conditioned World Model}\label{sec:morph_wm}
To effectively utilize the morphology vector $\mu$, we modify the standard RSSM at two points, the recurrent model (\autoref{eq:rnn}) and the observation encoder feeding the representation model (\autoref{eq:post}). A naive concatenation of static features $\mu$ with high-frequency observations $o_t$ often leads to the static signal being ``washed out'' by the high-variance dynamic data. To mitigate this, we employ a dual-tower encoder architecture in which the observation encoder is split into two pathways. The \textit{dynamic tower} processes the proprioceptive stream $o_t$ via an MLP. Simultaneously, the \textit{static tower} processes the morphology vector $\mu$ via a separate MLP to project it into the same latent dimension. These are concatenated and fused into the observation embedding $e_t$ consumed by the representation model (\autoref{eq:post}) as $e_t = \text{Linear}([\text{MLP}_{\text{dyn}}(o_t), \text{MLP}_{\text{stat}}(\mu)])$. The \ourMethod posterior is therefore $z_t \sim q_\phi(z_t \mid h_t, e_t)$ with $e_t$ being a function of both $o_t$ and $\mu$, so the posterior has access to $\mu$ through the static tower. This ensures that the encoder allocates dedicated capacity to representing the robot's physical structure before fusing it with the state.

Notably, the morphology $\mu$ is not just used for encoding. It is also injected directly into the recurrent dynamics. We augment \autoref{eq:rnn}'s transition function so that it explicitly depends on the robot type:
\begin{align}
    h_t = f_\phi(h_{t-1}, z_{t-1}, a_{t-1}, \mu)
\end{align}
By providing $\mu$ to the recurrent model at every time step, we relieve it of the burden of memorizing static physical properties in its short-term memory $h_t$. This disentanglement allows the recurrent state to focus purely on the \textit{dynamic} state of the system (e.g., velocity, contact timing) while the explicit conditioning handles the \textit{static} physics (e.g., limb length, mass), facilitating robust control across distinct morphologies.

\subsection{Physical Morphology Encoder}\label{sec:morph_embodiment}
Rather than treating morphology as a variable inferred from interaction history \citep{kumar2021rma} or an imperfect description, we formalize it as a mapping $\Phi: \mathcal{U} \rightarrow \mathbb{R}^d$ from the robot's description space $\mathcal{U}$ to a raw feature vector $\mu_\text{raw}$, which we then normalize into the morphology vector $\mu$. Following prior work \citep{chen2018hardware, mishra2025mcarl, xiong2023universal}, we extract features that parameterize the equations of motion, decomposing the description into kinematic, geometric, and dynamic properties that capture variations in leverage, stability margins, and rotational inertia. $\mu$ assumes a quadrupedal template with three-segment legs, parameterizing variation \textit{within} this family. We decompose it into four groups as follows.





\textit{Kinematics \& Topology} ($\mu_{\text{kin}}$) extracts segment lengths and knee configuration by traversing the front-left leg \citep{wang2018nervenet, huang2020one}, distinguishing inward ``X'' knees from outward ``dog-like'' configurations (\autoref{fig:overall_quads}). \textit{Geometry} ($\mu_{\text{geo}}$) describes the robot's footprint via the longitudinal and lateral distances between hip joints together with their aspect ratio, a scale-invariant indicator of body shape that ranges from a narrow lateral stance to a square footprint. \textit{Dynamics} ($\mu_{\text{dyn}}$) captures mass distribution, which governs rotational inertia and force response, and is encoded as a log-scaled total mass alongside the trunk mass fraction. Finally, \textit{Actuation} ($\mu_{\text{act}}$) captures weight-normalized torque to inform cross-embodiment transfer \citep{hwangbo2019learning, kumar2021rma}, accounting for how sufficient torque scales depending on the robot's weight.

The vector $\mu_\text{raw} = [\mu_\text{kin}, \mu_\text{geo}, \mu_\text{dyn}, \mu_\text{act}]$ mixes various units, so we apply min-max normalization to obtain $\mu = \Psi(\mu_\text{raw}) \in [-1, 1]^d$. Importantly, $\mu$ encodes \textit{idealized} rigid-body properties, and unmodeled residuals such as payload variation are absorbed instead by the dynamic latents $(h_t, z_t)$. These let the model distinguish, for example, the high-center-of-mass Spot from the compact Go1 (\aref{app:morphology}).

\subsection{Adaptive Reward Normalizer}\label{sec:hetero_rew}
Training a single agent across heterogeneous robots raises a reward-scaling challenge. All robots share the velocity-tracking objective, yet their reward formulations and penalty weights differ to accommodate distinct hardware (\aref{app:env_details}). Since the WM learns to predict rewards alongside dynamics, training on raw rewards causes robots with larger rewards to unevenly dominate the loss.

We therefore implement ARN using quantile-based scaling. Rather than a fixed scalar, we maintain per robot-type an EMA (momentum $\alpha$) of the inter-percentile range between the $\nth{5}$ and $\nth{95}$ percentiles of the returns, which yields the reward scale $\sigma_R$. The $\operatorname{max}$ in the denominator prevents the inflation of trivially small rewards early in training and avoids division by zero:
{\subfootnotesize
\begin{align}
    \sigma_R\!\leftarrow\!\alpha\sigma_R+(1-\alpha)(P_{95}-P_{05}),\quad r_{\text{norm}}\!=\!r/\max(1.0,\sigma_R)
\end{align}}
The mechanism acts as an adaptive equalizer. When a robot consistently generates large torque penalties, its $\sigma_R$ expands and dynamically scales the signal down, so \ourMethod receives a learning signal comparable in scale across morphologies. This prevents any single embodiment from dominating the latent representation and enables stable, simultaneous multi-robot training.

\section{Experiments}\label{sec:exps}
Our experiments test the claim that given the same morphology specification $\mu$, conditioning a learned dynamics model on $\mu$ and extracting a policy in imagination generalizes to unseen morphologies better than conditioning a model-free policy on the same $\mu$. We organize the evidence as follows. Both routes master the heterogeneous training cohort, and the model-free route reaches the same performance with more data, so any later gap is a generalization gap and not a capability gap (\autoref{sec:morph_mastery}). \ourMethod reproduces the dynamics of diverse robots over long horizons, which is the substrate the model-based route relies on (\autoref{sec:dynamics_fidelity}). On held-out morphologies, \ourMethod stays near specialist performance while a model-free policy given the identical $\mu$ degrades and collapses (\autoref{sec:generalization}). The frozen zero-shot policies then run on real hardware (\autoref{sec:realworld}). We also show that the learned latent state absorbs residuals when the specification is inaccurate or the robot drifts from it, which a pure specification-driven controller cannot (\aref{app:perturbation}).
We assume throughout that an accurate $\mu$ is available and that domain randomization spans the deployment conditions, so the residuals between the idealized $\mu$ and the real robot stay within the range the latent state has learned to absorb. Unless noted, we train on the full heterogeneous cohort.

\begin{figure*}[t]
    \centering
    \includegraphics[width=\textwidth]{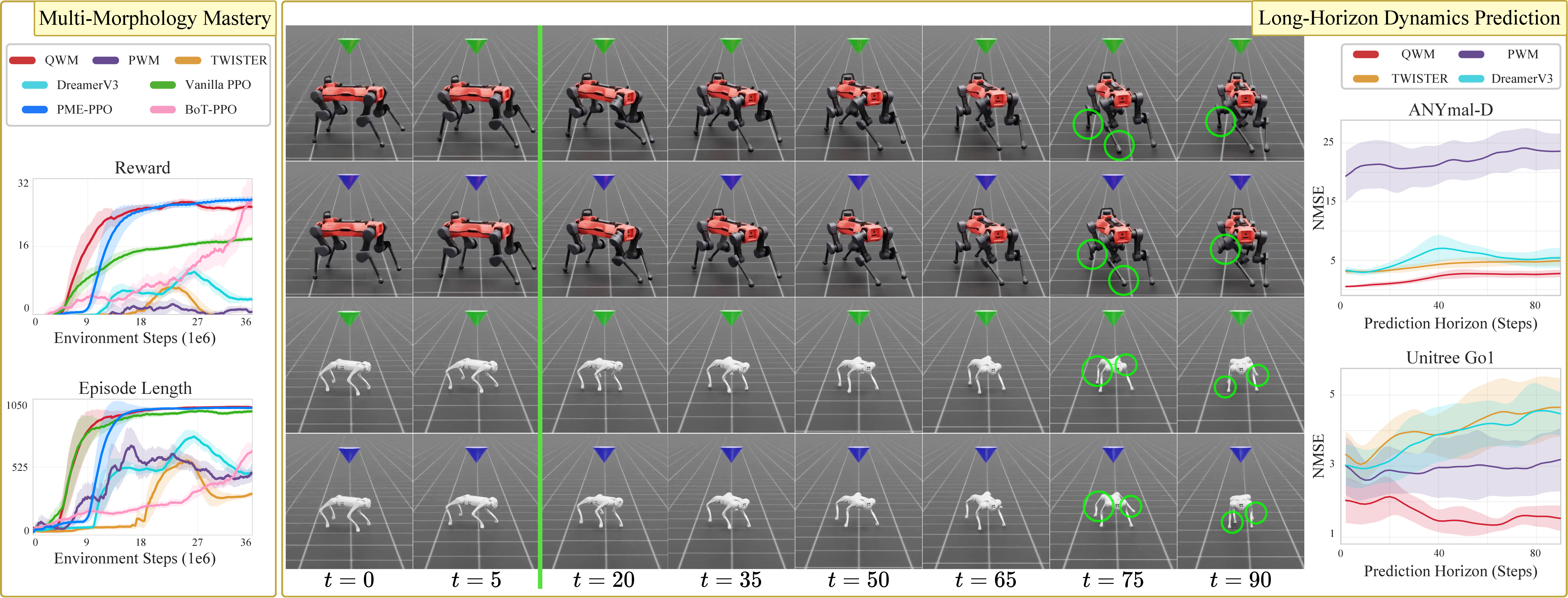}
    \caption{Training performance and dynamics fidelity on the heterogeneous cohort. Shaded regions denote std over 5 seeds.
    \textbf{Left panel}: Learning curves on the full heterogeneous cohort, with mean reward (top) and mean episode length (bottom).
    \textbf{Right panel}: Long-horizon dynamics prediction. Open-loop \textcolor[HTML]{048004}{ground truth} is compared against \textcolor[HTML]{0404d6}{imagination} over 85 steps, where \ourMethod stays synchronized across scales with minor late deviations (circled). The plots report NMSE over the training prediction window (32 trajectories), normalized by each robot's natural motion variance, where \ourMethod accumulates error gradually but does not diverge.}
\label{fig:perf_and_horizon}
\end{figure*}

\subsection{Multi-Morphology Mastery}\label{sec:morph_mastery}
We first test whether \ourMethod handles the interference that arises from training across morphologies simultaneously. We compare against three model-free baselines and three WM methods. The model-free baselines are Vanilla PPO \citep{schulman2017proximal}, PME-PPO (our PME conditioning $\mu$ into PPO via concatenation), and Body Transformer (BoT) \citep{sferrazza2024body}, which models the body as a graph via masked attention. The WM baselines are DreamerV3 \citep{hafner2025mastering}, PWM \citep{georgiev2025pwm}, and TWISTER \citep{burchi2025learning}, all of which must infer morphology implicitly from proprioceptive history without our PME or ARN. We track mean reward (\aref{app:reward_weights}) and episode length (up to 1000 steps) across 4,096 parallel environments.

As shown in \autoref{fig:perf_and_horizon} (left panel), the methods that receive $\mu$ explicitly master the cohort, while the methods that must infer it from history do not. Among the model-free baselines, PME-PPO reaches the same asymptotic performance as \ourMethod and clearly exceeds Vanilla PPO, although it needs roughly twice the environment steps to begin converging. BoT-PPO trails PME-PPO, since its high-capacity graph attention carries more optimization and sample cost than our compact MLP encoder, which makes PME-PPO the strongest like-for-like model-free baseline rather than a weak one. The WM baselines, which infer morphology from proprioceptive history rather than from an explicit $\mu$, settle into a mean-dynamics solution on this heterogeneous cohort and do not reach a controllable policy. We read this not as a verdict on implicit identification, which targets the harder setting where no description is available, but as motivation for our premise. When an accurate description exists, supplying $\mu$ explicitly removes the need to recover it online, and the open question is how to use it. The training curves settle the first half of that question. Both PME-PPO and \ourMethod, given the identical $\mu$, learn the full cohort, so $\mu$ is a sufficient statistic for mastering these kinematic chains under either learning paradigm. They differ only in cost, with \ourMethod reaching competitive reward and episode length in about half the environment steps of PME-PPO and additionally yielding a reusable dynamics model. On the training cohort the two routes are therefore comparable in capability, which isolates the model-free versus model-based question for the generalization experiments that follow.

\textbf{Component ablations.} \aref{app:ablation} localizes these gains. ARN is a precondition because every variant without it flatlines near zero reward, confirming that unnormalized heterogeneous reward scales alone prevent the WM from forming. Given ARN, the two morphology-injection sites play distinct, dissociable roles. Removing encoder conditioning permits early learning but triggers a late catastrophic collapse, indicating the latent posterior must be grounded in static physical context to remain stable over long horizons. Removing RSSM conditioning instead degrades sample efficiency and leaves a sub-optimal asymptotic reward, showing the recurrent state cannot retain morphology without step-wise re-injection. Removing explicit morphology entirely still learns to stay upright but plateaus below the full method, so implicit identification suffices for basic locomotion but not for refined gaits. Each component is necessary, and they fail in different ways.

\subsection{Long-Horizon Dynamics Prediction}\label{sec:dynamics_fidelity}

To verify that \ourMethod functions as a reliable neural simulator, we evaluate its open-loop prediction fidelity. In \autoref{fig:perf_and_horizon} (right panel), we initialize with a short context ($T=5$) and unroll 85 steps of pure imagination on ANYmal-D and Unitree Go1, feeding only the policy's actions. The long horizon exposes gait-level divergence that shorter rollouts would hide. The imagined trajectories stay tightly synchronized with the ground truth, with only minor deviations near the end of the rollout, confirming that \ourMethod parameterizes dynamics for diverse robots simultaneously.

Quantitatively (\autoref{fig:perf_and_horizon} (right panel-plots)), we report the Normalized Mean Squared Error (NMSE) of predicted proprioception over the training prediction window (32 episodes per robot, normalized by per-robot motion variance for cross-morphology comparison). \ourMethod maintains the lowest error among the learned WMs with minimal accumulation, which indicates a stable physical attractor rather than drift. The models that infer morphology implicitly do not reach this fidelity on the heterogeneous cohort, consistent with their lower training performance. Without explicit conditioning, a single WM must average over the conflicting dynamics of a diverse cohort, which is the failure mode that our $\mu$ injection is designed to remove.

\subsection{Zero-Shot Generalization to Unseen Morphologies}\label{sec:generalization}
To test whether \ourMethod learns physics-grounded dynamics rather than overfitting to specific hardware, we evaluate on held-out morphologies under two regimes drawn from our per-feature leave-one-out (LOO) deviation analysis (\aref{app:morphology}), one holding out robots whose every feature lies within the cohort's range (Go1, ANYmal-D) and one holding out a robot that deviates strongly on several axes at once (B2). To evaluate a target robot $r$, we train models from scratch on the seven robots excluding $r$. We characterize each case by its full per-feature deviation profile rather than a single scalar distance, since features affect dynamics non-uniformly.

\begin{wraptable}{r}{0.5\textwidth}
\centering
\scriptsize
\renewcommand{\arraystretch}{0.8}
\setlength{\tabcolsep}{2.5pt}
\caption{Zero-shot generalization performance on unseen robots. Mean $\pm$ std over 5 seeds (128 envs/seed).}
\begin{tabular}{@{}l l c c c@{}}
\toprule
\textbf{Method} & \textbf{Metric} & \textbf{ANYmal-D} & \textbf{Unitree Go1} &
\textbf{Unitree B2} \\
\midrule
\multirow{2}{*}{PME-PPO} & Rwd. & $10.1 \pm 1.9$ & $23.1 \pm 2.3$ &
$-0.2 \pm 3.2$ \\
 & Len. & $530 \pm 8.2$ & $602 \pm 14.9$ &
$337 \pm 23.1$ \\
\midrule
\multirow{2}{*}{\ourMethod} & Rwd. & $18.2 \pm 2.3$ & $35.5 \pm 3.4$ &
$12.1 \pm 4.2$ \\
 & Len. & $948.6 \pm 12.1$ & $974.4 \pm 6.2$ &
$925.4 \pm 17.2$ \\
\midrule
Specialist & Rwd. & $21.8 \pm 1.2$ & $39.7 \pm 2.1$ &
$13.6 \pm 4.2$ \\
PPO & Len. & $981.3 \pm 4.2$ & $996.1 \pm 1.1$ &
$961.1 \pm 8.9$ \\
\bottomrule
\end{tabular}
\label{tab:zero_shot_results}
\end{wraptable}

In all cases, all weights are frozen and conditioned solely on the target's $\mu$. Our central comparison is against PME-PPO, the model-free policy given the same $\mu$, which isolates whether generalization stems from $\mu$ alone or from the learned dynamics, with a Specialist PPO oracle trained per target robot as an empirical upper bound. We omit the implicit baselines, which did not reach a controllable policy on the training cohort (\autoref{sec:morph_mastery}).

\textbf{Within-range transfer.} On the training cohort the two routes are comparable (\autoref{fig:perf_and_horizon} (left panel)), so any gap here is a generalization gap rather than a capability one. On the held-out Go1 and ANYmal-D (\autoref{tab:zero_shot_results}), \ourMethod matches specialist episode length and recovers 80-90\% of specialist reward, while PME-PPO, given the same $\mu$ but no learned dynamics, trails substantially (episode length 602 vs. 974 on Go1). Both robots sit inside the cohort's per-axis support (\aref{app:morphology}). Because each axis is individually represented in training, the WM synthesizes the held-out dynamics by interpolating its learned latent geometry along each physical dimension.

\textbf{Out-of-range transfer.} Under the same comparison, B2 is elevated on several axes at once, all pushing the same direction toward larger, heavier, and stronger. Each deviation is individually bounded, but no single training robot combines this inertia with this stance and actuator capacity, leaving the configuration jointly unrepresented. This is a combinatorial gap rather than a large scalar distance, which is why a model-free policy given $\mu$ cannot bridge it. \ourMethod, on the same $\mu$, instead reaches near-specialist locomotion zero-shot, within about 4\% of the episode-length and 11\% of the reward upper bound. The $\mu$-conditioned latent dynamics thus supply the inductive bias to synthesize a controller for a jointly unseen configuration, combining mild extrapolation on mass and torque capacity with interpolation along the remaining axes rather than relying on a single nearest neighbor. Latent-space evidence, that kinematically similar robots occupy neighboring regions while dynamic state varies smoothly within each, appears in \aref{app:latent_viz}.

The gain comes from learned dynamics, not from access to $\mu$. Two perturbation studies isolate this (\aref{app:perturbation}). When we corrupt $\mu$ on a nominal world, \ourMethod retains $\sim$80\% of its episode length even at 20\% corruption while PME-PPO, reading the same $\mu$, falls to roughly a third, because \ourMethod's latent state $z_t$ re-reads the true dynamics from observations and corrects the mis-specified input. When we instead hold $\mu$ correct and perturb the true world through torque loss or added payload, even a near-exact analytical rigid-body predictor conditioned on the nominal $\mu$ diverges, since it simulates a robot that no longer exists, whereas \ourMethod stays bounded and retains 75-84\% of its nominal episode length. A model-free $\mu$-policy and a static analytical prior both lack this correction channel, so the advantage we report is a property of routing $\mu$ through learned dynamics rather than holding $\mu$ at all.

\subsection{Real-World Deployment}\label{sec:realworld}

\begin{wrapfigure}{r}{0.43\textwidth}
  \centering
  \begingroup
  \captionsetup{type=table}
  \scriptsize
  \renewcommand{\arraystretch}{0.8}
  \setlength{\tabcolsep}{2.5pt}
  \caption{Quantitative real-world results (10 trials $\times$ 60s per platform).}
  \label{tab:realworld}
  \begin{tabular}{@{}ll cc}
  \toprule
  \textbf{Platform} & \textbf{Method} & $e_{xy}$ (m/s) $\downarrow$ & $e_{yaw}$ (rad/s) $\downarrow$ \\
  \midrule
  \multirow{2}{*}{ANYmal-D}
    & \ourMethod & $0.30$ & $0.29$ \\
    & Specialist PPO         & $0.28$ & $0.26$ \\
  \midrule
  \multirow{2}{*}{Unitree Go1}
    & \ourMethod & $0.34$ & $0.34$ \\
    & Specialist PPO         & $0.31$ & $0.30$ \\
  \bottomrule
  \end{tabular}
  \endgroup

  \vspace{1.em}

  \includegraphics[width=0.8\linewidth]{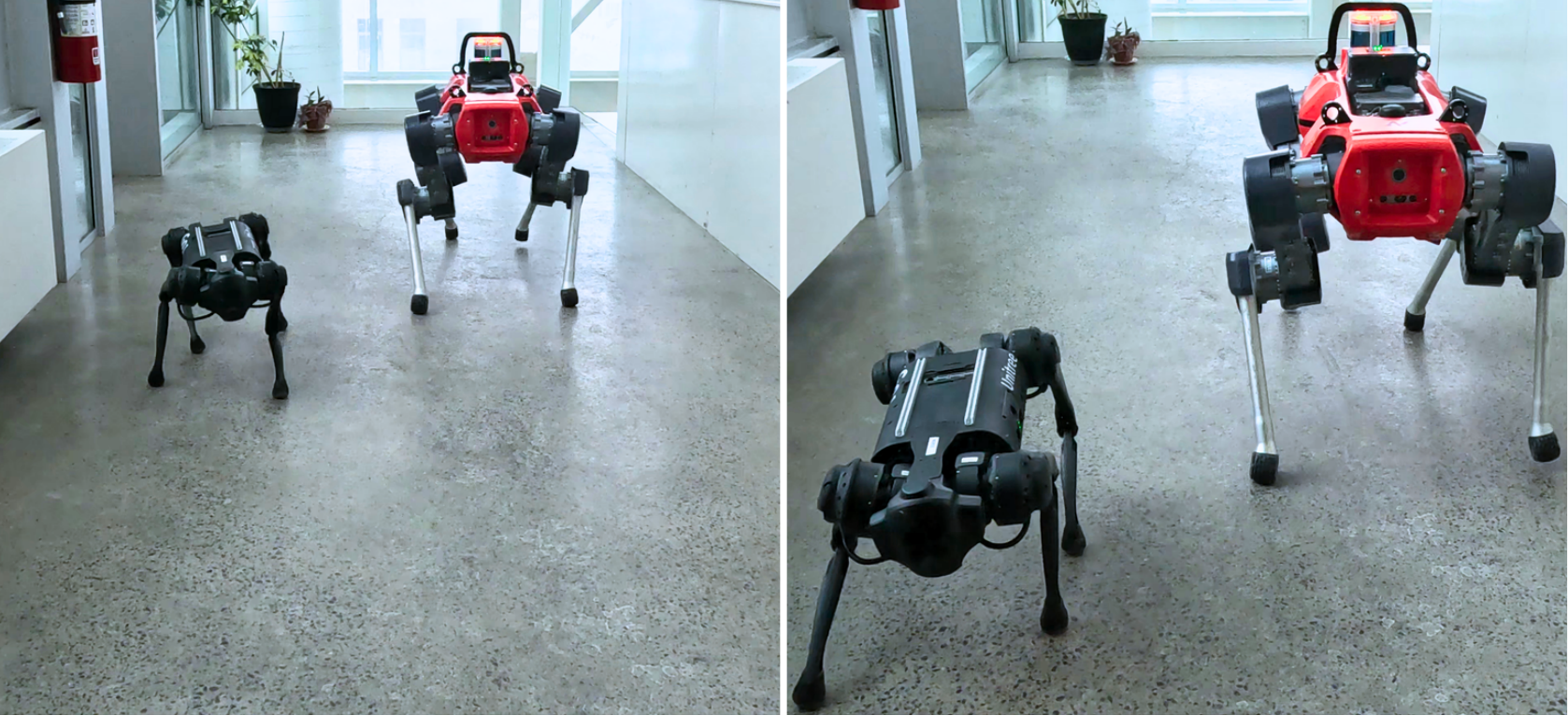}
  \caption{Real-world deployment on the held-out Unitree Go1 and ANYmal-D.}
  \label{fig:deployment}

\end{wrapfigure}
The physical world introduces unmodeled dynamics absent from both the robot specification and simulation, such as actuator backlash and motor nonlinearities. We deploy the two frozen zero-shot agents from \autoref{sec:generalization}, the held-out ANYmal-D and Go1, directly onto hardware using the exact simulation policy weights, with inference at 50Hz on each platform's onboard computer (\autoref{fig:deployment}). This zero-shot behavior transferred directly to hardware. The agent produced a high-frequency trot on the agile Go1 and a slower, grounded gait on the heavier ANYmal-D, achieving stable locomotion on both without any real-world fine-tuning. As shown in \autoref{tab:realworld}, \ourMethod tracks velocity commands within $\sim$10-13\% of the specialist baseline with zero falls across all trials, despite never interacting with either robot during training, confirming that explicit morphology conditioning encodes the physics needed to control unseen real robots rather than simulation artifacts. Further details in \aref{app:realworld}.

\section{Limitations}
Our results are bounded in scope. The vector $\mu$ encodes a fixed quadrupedal template with three-segment legs, a 12-DoF action space, and a stance rooted at the hip joints, so generalization is parametric \textit{within} this template and not across structural variants such as differing joints per leg or non-quadrupedal bodies. Our evaluation is limited to ``blind'' flat-ground velocity tracking without exteroception or contact-rich terrain.

\section{Conclusion}
We introduce \ourMethod, a step toward generalist dynamics model achieving zero-shot sim-to-real locomotion across diverse quadrupeds by conditioning on scale-invariant physical features. Crucially, these learned dynamics absorb real-world physical divergences, a capability we validate against a model-free controller that fails on outliers despite receiving the identical $\mu$. Ultimately, \ourMethod shows that cross-embodiment morphology data must route through learned dynamics rather than directly into policies, as dynamics within a morphological family vary smoothly unlike optimal policies.

\bibliography{references}

\clearpage
\appendix
\include{supplementary_v2}

\end{document}

%% file: math_commands.tex
\usepackage{amsmath,amsfonts,bm}

\usepackage{xspace}

\def\eqref#1{equation~\ref{#1}}

\def\1{\bm{1}}

\DeclareMathAlphabet{\mathsfit}{\encodingdefault}{\sfdefault}{m}{sl}
\SetMathAlphabet{\mathsfit}{bold}{\encodingdefault}{\sfdefault}{bx}{n}



%% file: supplementary_v2.tex
\section{Environment Details \& Reward Structure}
\label{app:env_details}

In this section, we provide the detailed specification of the environment, reward structures, and randomization parameters used to train \ourMethod.

To demonstrate the robustness of our architecture and ensure fair comparison against baselines, we purposefully avoided extensive hyperparameter tuning. Instead, we adopted the default reward terms, weights, and randomization ranges provided by the official NVIDIA Isaac Lab repository \citep{mittal2025isaaclab} for each respective robot asset. By relying on these standard, community-vetted configurations, we show that \ourMethod can generalize across morphologies without requiring the reward landscape to be engineered specifically for our latent representation.

\subsection{Reward Functions}
The reward definitions vary between the standard quadruped families (ANYmal and Unitree series) and the Boston Dynamics Spot, reflecting the different control strategies often employed for industrial versus agile platforms. We denote the robot state available for reward computation by linear velocity $\mathbf{v} \in \mathbb{R}^3$, angular velocity $\boldsymbol{\omega} \in \mathbb{R}^3$, joint positions $\mathbf{q} \in \mathbb{R}^{12}$, joint velocities $\dot{\mathbf{q}} \in \mathbb{R}^{12}$, joint torques $\boldsymbol{\tau} \in \mathbb{R}^{12}$, and the projected gravity vector $\mathbf{g}_{proj} \in \mathbb{R}^3$. The agent's action (target joint positions) is denoted $\mathbf{a}_t \in \mathbb{R}^{12}$. The commanded velocities are $\mathbf{v}^{cmd}_{xy} \in \mathbb{R}^2$ and $\omega^{cmd}_z \in \mathbb{R}$ (or heading commands where applicable).

The policy observation space ($\mathbb{R}^{48}$) consists of $\mathbf{v}$, $\boldsymbol{\omega}$, $\mathbf{g}_{proj}$, $\mathbf{v}^{cmd}$, $\mathbf{q} - \mathbf{q}_{default}$, $\dot{\mathbf{q}}$, and the previous actions $\mathbf{a}_{t-1}$.

For the ANYmal (B, C, D) and Unitree (A1, Go1, Go2, B2) families, we utilize the standard ``Flat Terrain'' reward structure common in robotic learning \citep{mittal2025isaaclab}. As shown in \autoref{tab:standard_rewards}, the objective is primarily velocity tracking with auxiliary penalties to encourage energy efficiency and smoothness. We employ the squared exponential kernel $e^{-\|x\|^2/\sigma^2}$ for tracking terms.

\begin{table}[ht]
\centering
\caption{Standard Reward Functions (ANYmal \& Unitree Families)}
\label{tab:standard_rewards}
\small
\setlength{\tabcolsep}{4pt}
\renewcommand{\arraystretch}{1.2}
\begin{tabular}{ll|ll}
\toprule
\textbf{Tracking Term} & \textbf{Equation} & \textbf{Penalty Term} & \textbf{Equation} \\
\midrule
Lin. Vel. Track (XY) & $e^{-\|\mathbf{v}_{xy} - \mathbf{v}^{cmd}_{xy}\|^2 / \sigma_{v}^2}$ & Lin. Vel. Z & $- v_z^2$ \\
Ang. Vel. Track (Z) & $e^{-(\omega_z - \omega^{cmd}_z)^2 / \sigma_{\omega}^2}$ & Ang. Vel. XY & $- \|\boldsymbol{\omega}_{xy}\|^2$ \\
Feet Air Time & $\sum (t_{air} - 0.5) \cdot \mathbb{I}_{landed}$ & Flat Orientation & $- \|\mathbf{g}_{proj, xy}\|^2$ \\
& & Torque & $- \|\boldsymbol{\tau}\|^2$ \\
& & Joint Acc. & $- \|\ddot{\mathbf{q}}\|^2$ \\
& & Action Rate & $- \|\mathbf{a}_t - \mathbf{a}_{t-1}\|^2$ \\
& & Contacts\textsuperscript{*} & $- \mathbb{I}(\|\mathbf{F}_{c}\| > 1.0)$ \\
\bottomrule
\multicolumn{4}{l}{\footnotesize \textsuperscript{*}Enabled for the ANYmal series and disabled for the Unitree series.}
\end{tabular}
\end{table}

The Boston Dynamics Spot utilizes a distinct reward setup derived from its specific configuration in Isaac Lab. Unlike the standard group, this configuration includes explicit ``gait enforcement'' terms (\autoref{tab:spot_rewards}) to produce a specific trotting style, along with stricter kinematic constraints (e.g., foot clearance).

\begin{table}[ht]
\centering
\caption{Spot-Specific Reward Functions}
\label{tab:spot_rewards}
\small
\setlength{\tabcolsep}{3pt}
\renewcommand{\arraystretch}{1.3}
\begin{tabular}{ll|ll}
\toprule
\textbf{Functional Term} & \textbf{Equation} & \textbf{Penalty Term} & \textbf{Equation} \\
\midrule
Base Lin. Vel (Ramp) & $m \cdot e^{-\|\mathbf{v}_{xy} - \mathbf{v}^{cmd}_{xy}\|/\sigma_v}$ & Foot Slip & $- \sum \|\mathbf{v}_{f, xy}\| \cdot \mathbb{I}_{contact}$ \\
\multicolumn{1}{r}{\textit{where } $m =$} & $1 + 0.5(\|\mathbf{v}^{cmd}_{xy}\| - 1)_{+}$ & Base Motion & $- (0.8 v_z^2 + 0.2 \|\boldsymbol{\omega}_{xy}\|_1)$ \\
Base Ang. Vel (Abs) & $e^{-|\omega_z - \omega^{cmd}_z| / \sigma_{\omega}}$ & Base Orientation & $- \|\mathbf{g}_{proj, xy}\|$ \\
Foot Clearance & $e^{-\sum (h_f - 0.1)^2 \tanh(2 v_{f})}$ & Act. Smoothness & $- \|\mathbf{a}_t - \mathbf{a}_{t-1}\|$ \\
Gait Enforcement & $r_{sync} \cdot r_{async}$ & Air Time Var. & $- (\text{Var}(t_{air}) + \text{Var}(t_{cont}))$ \\
\multicolumn{1}{r}{\textit{where } $r_i =$} & $e^{-(\Delta t_{air}^2 + \Delta t_{cont}^2)/\sigma}$ & Joint Pos/Vel & $- \|\mathbf{q} - \mathbf{q}_{0}\|, - \|\dot{\mathbf{q}}\|$ \\
& & Contacts & $- \mathbb{I}(\|\mathbf{F}_{c}\| > 1.0)$ \\
\bottomrule
\end{tabular}
\end{table}

\subsection{Reward Weights}
\label{app:reward_weights}
\autoref{tab:reward_weights} and \autoref{tab:spot_weights} detail the coefficients used for scalarization. We preserve the heterogeneous weights from the optimized single-robot configurations rather than normalizing them manually.

\begin{table}[ht]
\centering
\caption{Reward Weights for ANYmal and Unitree Robots}
\label{tab:reward_weights}
\resizebox{\textwidth}{!}{
\begin{tabular}{l|ccc|cccc}
\toprule
\textbf{Reward Term} & \textbf{ANYmal-D} & \textbf{ANYmal-C} & \textbf{ANYmal-B} & \textbf{Unitree A1} & \textbf{Unitree Go1} & \textbf{Unitree Go2} & \textbf{Unitree B2} \\
\midrule
Lin. Vel. Tracking (XY) & 1.0 & 1.0 & 1.0 & 1.5 & 1.5 & 1.5 & 1.0 \\
Ang. Vel. Tracking (Z) & 0.5 & 0.5 & 0.5 & 0.75 & 0.75 & 0.75 & 0.75 \\
Feet Air Time & 0.5 & 0.5 & 0.5 & 0.25 & 0.25 & 0.25 & 0.5 \\
\midrule
Lin. Vel. Z ($L^2$) & -2.0 & -2.0 & -2.0 & -2.0 & -2.0 & -2.0 & -2.0 \\
Ang. Vel. XY ($L^2$) & -0.05 & -0.05 & -0.05 & -0.05 & -0.05 & -0.05 & -0.05 \\
Flat Orientation ($L^2$) & -5.0 & -5.0 & -5.0 & -2.5 & -2.5 & -2.5 & -5.0 \\
Joint Torques ($L^2$) & -2.5e-5 & -2.5e-5 & -2.5e-5 & -2.0e-4 & -2.0e-4 & -2.0e-4 & -2.0e-5 \\
Joint Acc. ($L^2$) & -2.5e-7 & -2.5e-7 & -2.5e-7 & -2.5e-7 & -2.5e-7 & -2.5e-7 & -2.5e-7 \\
Action Rate ($L^2$) & -0.01 & -0.01 & -0.01 & -0.01 & -0.01 & -0.01 & -0.01 \\
Undesired Contacts & -1.0 & -1.0 & -1.0 & - & - & - & - \\
\bottomrule
\end{tabular}
}
\end{table}

\begin{table}[ht]
\centering
\caption{Reward Weights for Boston Dynamics Spot}
\label{tab:spot_weights}
\begin{tabular}{lc|lc}
\toprule
\textbf{Reward Term} & \textbf{Weight} & \textbf{Reward Term} & \textbf{Weight} \\ \midrule
Gait Reward & 10.0 & Base Motion Penalty ($v_z, \boldsymbol{\omega}_{xy}$) & -2.0 \\
Base Linear Vel (Ramp) & 5.0 & Base Orientation Penalty & -3.0 \\
Base Angular Vel & 5.0 & Action Smoothness & -1.0 \\
Air Time Reward & 5.0 & Air Time Variance & -1.0 \\
Foot Clearance & 0.5 & Foot Slip & -0.5 \\
Joint Position Penalty & -0.7 & Joint Acceleration & -1.0e-4 \\
Joint Velocity Penalty & -0.01 & Joint Torques & -5.0e-4 \\
Undesired Contacts & -1.0 & & \\
\bottomrule
\end{tabular}
\end{table}

\subsection{Domain Randomization}
To ensure sim-to-real transfer, we apply extensive domain randomization.

\subsubsection{Physics \& Observation Randomization}
We apply standard randomization to surface friction and observation noise (\autoref{tab:phys_rand}).

\begin{table}[ht]
\centering
\caption{Physics and Observation Randomization Parameters}
\label{tab:phys_rand}
\begin{tabular}{l|l}
\toprule
\textbf{Parameter} & \textbf{Value / Distribution} \\
\midrule
Static Friction & $\mathcal{U}(0.8, 0.8)$ (Default) \\
Dynamic Friction & $\mathcal{U}(0.6, 0.6)$ (Default) \\
External Disturbance & $v_{push} \sim \mathcal{U}(-0.5, 0.5)$ m/s (Every 10 to 15s) \\
Control Latency & 20ms (Simulation $dt=0.005$s, Decimation=4) \\
\midrule
\textit{Observation Noise} & \textit{Additive Uniform Distribution} \\
Linear Velocity & $\mathcal{U}(-0.1, 0.1)$ m/s \\
Angular Velocity & $\mathcal{U}(-0.2, 0.2)$ rad/s \\
Projected Gravity & $\mathcal{U}(-0.05, 0.05)$ \\
Joint Positions & $\mathcal{U}(-0.01, 0.01)$ rad \\
Joint Velocities & $\mathcal{U}(-1.5, 1.5)$ rad/s \\
\bottomrule
\end{tabular}
\end{table}

\subsubsection{Morphology Randomization}
We randomize physical properties during training. We define two randomization groups, Large (ANYmal, B2, Spot) and Small (A1, Go1, Go2), to ensure that mass perturbations are relative to the robot's size.

\begin{table}[ht]
\centering
\caption{Morphology Randomization Parameters}
\label{tab:morph_rand}
\begin{tabular}{l|l}
\toprule
\textbf{Parameter} & \textbf{Distribution / Range} \\
\midrule
Mass (Large Robots) & Base $\pm 5.0$ kg \\
Mass (Small Robots) & Base $+ [-1.0, 3.0]$ kg \\
\midrule
Center of Mass (Base) & $\pm 5$cm (X, Y), $\pm 1$cm (Z) \\
\bottomrule
\end{tabular}
\end{table}

\subsubsection{Reset Randomization \& Task Commands}
We utilize different initialization strategies based on robot sensitivity (\autoref{tab:reset_rand}). While larger robots utilize joint scaling, smaller Unitree platforms are initialized at nominal joint positions to ensure stability. The velocity command ranges (\autoref{tab:commands}) are uniform across all robots in the heterogeneous setup.

\begin{table}[ht]
\centering
\caption{Reset Randomization}
\label{tab:reset_rand}
\begin{tabular}{l|l}
\toprule
\textbf{State} & \textbf{Initialization} \\
\midrule
Base Position & Offset $\pm 0.5$m (X,Y) \\
Base Yaw & $\mathcal{U}(-\pi, \pi)$ \\
Base Velocity (Lin/Ang) & $\mathcal{U}(-0.5, 0.5)$ (Zero for Small Unitrees) \\
Joint Pos (ANYmal, B2) & Nominal $\times \mathcal{U}(0.8, 1.2)$ \\
Joint Pos (Small Unitree) & Nominal (Fixed) \\
Joint Pos (Spot) & Nominal $+ \mathcal{U}(-0.05, 0.05)$ rad \\
\bottomrule
\end{tabular}
\end{table}

\begin{table}[ht]
\centering
\caption{Velocity Command Ranges}
\label{tab:commands}
\begin{tabular}{l|ccc}
\toprule
\textbf{Robot Group} & $\mathbf{v}_x$ (m/s) & $\mathbf{v}_y$ (m/s) & $\boldsymbol{\omega}_z$ (rad/s) \\
\midrule
All Robots & $[-1.0, 1.0]$ & $[-1.0, 1.0]$ & $[-1.0, 1.0]$ \\
\bottomrule
\end{tabular}
\end{table}

\clearpage
\section{Heterogeneous Simulation Infrastructure}
We utilize NVIDIA Isaac Lab \citep{mittal2025isaaclab}, built upon Isaac Sim \citep{NVIDIA_Isaac_Sim}, as our simulation backbone. While Isaac Lab offers massive parallelism via GPU-accelerated PhysX, its default vectorization paradigm is designed for \textit{homogeneous} setups, spawning thousands of identical clones of a single robot asset. This standard setup is insufficient for training a generalizable agent across multiple robots \textit{simultaneously}, as it typically restricts the training loop to a single morphology per run.

To overcome this limitation, we developed a heterogeneous environment on top of Isaac Lab, \texttt{Hetero-Isaac}.

Unlike standard vectorization, \texttt{Hetero-Isaac} manages a heterogeneous batch where the robot type is indexed per environment ID, allowing the replay buffer to collect a mixed distribution of morphologies in a single forward pass. By managing unique robot description file paths and physics tensors, we construct a diverse morphology cohort including the ANYmal (B, C, D), Unitree (Go1, Go2, A1, B2), and Boston Dynamics Spot (\autoref{fig:overall_quads}). Importantly, this cohort is not merely a collection of scaled variants, but it captures fundamental kinematic shifts. The robots span distinct joint topologies, ranging from the ``X-configuration'' of the ANYmals (where knees bend inwards towards the chassis) to the standard mammalian ``dog-like'' configuration of the Spot and Unitree series. Furthermore, the hip-to-foot offsets and link length ratios vary non-linearly across the fleet. This structural diversity prevents the agent from relying on a single gait geometry, forcing the WM to capture the distinct underlying dynamics of each kinematic chain.

Essential to our setup, \texttt{Hetero-Isaac} ensures full physical fidelity by maintaining distinct collision geometries, kinematic trees, and actuator gains for each agent in parallel. Furthermore, it manages the complexity of \textit{heterogeneous reward definitions}. Different robots require distinct objective functions and weightings. For instance, Spot accumulates rewards on a vastly different scale ($\approx 350$ per episode) compared to the ANYmal-D ($\approx 25$) (see \autoref{sec:morph_embodiment}). \texttt{Hetero-Isaac} exposes these diverse, unnormalized signals directly, necessitating the adaptive per-robot reward normalizer described in \autoref{sec:hetero_rew}.

\textbf{Training distribution.} Throughout all experiments, \texttt{Hetero-Isaac} applies extensive domain randomization to expose policies to the kind of unmodeled phenomena present on real hardware, namely surface friction variation, mass perturbations, center-of-mass shifts, external push disturbances, control latency, and additive observation noise across all proprioceptive channels (full specification in \aref{app:env_details}). Consequently, all simulation results reported in this work, encompassing training curves, zero-shot transfer, and perturbation robustness, reflect performance across a \textit{distribution} of physical conditions per morphology, not a single nominal setting. This randomization is what allows the learned latent dynamics $(h_t, z_t)$ to absorb residuals beyond the idealized $\mu$ vector, and is a key reason \ourMethod transfers to real hardware without fine-tuning despite the sim-to-real gap.

\clearpage
\section{Network Architecture Details}
\label{app:network_details}

We build upon the DreamerV3 \citep{hafner2025mastering} backbone, introducing specific architectural modifications to handle heterogeneous morphologies. Unless otherwise specified, all dense layers are followed by \texttt{LayerNorm} and the \texttt{SiLU} activation function.

\subsection{Morphology-Conditioned Encoder}
To effectively fuse the high-frequency proprioceptive data ($o_t$) with the static, low-frequency morphology features ($\mu$), we employ a dual-pathway architecture, detailed in \autoref{sec:qwm}.

As seen in \autoref{box:encoder}, we process the modalities in separate pathways before concatenation. The proprioception tower is deep (5 layers) and wide (1024 units) to extract features from the complex joint state history. Conversely, the robot description tower is shallower (2 layers) and narrower (512 units) because the input features $\mu$ (ratios, normalized masses) are already semantically dense. This design prevents the static signal from being overwhelmed by the variance of the dynamic signal before they are projected into the shared latent space.

\begin{tcolorbox}[title={\textbf{Dual-Pathway Encoder}},
    colframe=blue!50!black,
    colback=blue!5!white,
    coltitle=white
]
\label{box:encoder}
\textbf{Inputs:}
\begin{itemize}
    \item Proprioception $o_t \in \mathbb{R}^{48}$ (Joint pos, vel, base vel, gravity, commands, actions)
    \item Morphology $\mu \in \mathbb{R}^{10}$ (Normalized robot description features)
\end{itemize}

\textbf{1. Proprioception Tower:}
\begin{verbatim}
# Wide and deep to capture high-freq dynamics
x_dyn = Linear(48, 1024) -> LayerNorm -> SiLU
x_dyn = Linear(1024, 1024) -> LayerNorm -> SiLU
x_dyn = Linear(1024, 1024) -> LayerNorm -> SiLU
x_dyn = Linear(1024, 1024) -> LayerNorm -> SiLU
x_dyn = Linear(1024, 1024) -> LayerNorm -> SiLU
\end{verbatim}

\textbf{2. Morphology Tower:}
\begin{verbatim}
# Shallower tower; features are already dense
# Units = 0.5 * Proprio_Units
x_stat = Linear(10, 512) -> LayerNorm -> SiLU
x_stat = Linear(512, 512) -> LayerNorm -> SiLU
\end{verbatim}

\textbf{3. Fusion Layer:}
\begin{verbatim}
concat = Concatenate([x_dyn, x_stat]) # dim: 1536
embed  = Linear(1536, 1024) -> LayerNorm -> SiLU
\end{verbatim}
\end{tcolorbox}

\subsection{Recurrent State-Space Model (RSSM)}
The core of the WM is the RSSM, which learns the transition dynamics $p(s_{t+1} | s_t, a_t, \mu)$. We utilize a discrete latent space (Categorical VAE) which has been shown to be more robust to world modeling errors than Gaussian equivalents.

Crucially, as shown in \autoref{box:rssm}, the morphology vector $\mu$ is injected again at the input of the Recurrent Cell. This explicit recurrent conditioning ensures that the GRU state $h_t$ does not need to memorize the robot's physical parameters (like limb length) in its short-term memory, freeing up capacity for tracking dynamic state (like velocity and contact timing).

\begin{tcolorbox}[
    title={\textbf{RSSM Dynamics}},
    colframe=blue!50!black,
    colback=blue!5!white,
    coltitle=white
]
\label{box:rssm}
\textbf{State Definitions:}
\begin{itemize}
    \item Deterministic State $h_t \in \mathbb{R}^{512}$ (GRU State)
    \item Stochastic State $z_t$: 32 categorical variables, 32 classes each ($32 \times 32$ flattened to 1024)
\end{itemize}

\textbf{1. Recurrent Step (Finding $h_t$):}
\begin{verbatim}
# Inputs: Previous stochastic state, Action, Morphology
# Note: z_{t-1} is flattened from (32, 32) -> 1024
input_rnn = Concatenate([Flatten(z_{t-1}), a_{t-1}, mu])

# Pre-processing MLP
x_rnn = Linear(input_rnn, 512) -> LayerNorm -> SiLU

# GRU Update
h_t = GRUCell(input=x_rnn, hidden_state=h_{t-1})
\end{verbatim}

\textbf{2. Posterior Estimation (Representation Model):}
\begin{verbatim}
# Fuses history (h_t) with current observation embedding
input_post = Concatenate([h_t, embed])
x_post = Linear(input_post, 512) -> LayerNorm -> SiLU
logits = Linear(512, 32 * 32)
z_t_post ~ OneHotCategorical(logits) # Used for training
\end{verbatim}

\textbf{3. Prior Estimation (Transition Predictor):}
\begin{verbatim}
# Predicts state purely from history (for imagination)
x_prior = Linear(h_t, 512) -> LayerNorm -> SiLU
logits = Linear(512, 32 * 32)
z_t_prior ~ OneHotCategorical(logits) # Used for planning
\end{verbatim}
\end{tcolorbox}

\subsection{Policy and Critic Heads}
The Actor and Critic networks operate on the concatenated state feature $feat_t = [h_t, \text{flatten}(z_t)]$.

The Critic utilizes \texttt{Symlog} distribution learning. Instead of regressing a single scalar value (which is unstable when rewards vary by orders of magnitude, as seen in \autoref{sec:hetero_rew}), the network predicts a categorical distribution over 255 bins. This allows the critic to model the multi-modal return distributions common in multi-task learning.

The Actor outputs a Gaussian distribution. To prevent premature convergence, the standard deviation is learned but bounded between $[0.1, 2.0]$.

\begin{tcolorbox}[
    title={\textbf{Actor \& Critic}},
    colframe=blue!50!black,
    colback=blue!5!white,
    coltitle=white
]
\label{box:heads}
\textbf{Input:} $feat_t = \text{Concatenate}([h_t, \text{Flatten}(z_t)])$ (dim: $512 + 1024 = 1536$)

\textbf{1. Actor (Policy):}
\begin{verbatim}
x = Linear(1536, 512) -> LayerNorm -> SiLU
x = Linear(512, 512) -> LayerNorm -> SiLU
x = Linear(512, 512) -> LayerNorm -> SiLU

# Mean head
mean = Linear(512, 12)

# Std head (Learned, constrained)
raw_std = Linear(512, 12)
std = Softplus(raw_std) + 0.1
std = clamp(std, min=0.1, max=2.0)

action ~ Normal(mean, std)
\end{verbatim}

\hrulefill

\textbf{2. Critic (Value Function):}
\begin{verbatim}
x = Linear(1536, 512) -> LayerNorm -> SiLU
x = Linear(512, 512) -> LayerNorm -> SiLU
x = Linear(512, 512) -> LayerNorm -> SiLU

# Symlog Discrete Output (TwoHot Encoding)
logits = Linear(512, 255)
value_dist = Softmax(logits)
# Expected value is calculated via weighted sum of bin centers
\end{verbatim}
\end{tcolorbox}

\subsection{Reward Head}
The reward predictor is structurally identical to the Critic, utilizing the same \texttt{Symlog} two-hot discretization. However, it is deeper (4 layers) than the policy heads. This increased depth is necessary because the reward function in locomotion is often a highly non-linear combination of the state (e.g., precise velocity tracking combined with sharp boolean penalties for collisions).

\begin{tcolorbox}[
    title={\textbf{Reward Predictor}},
    colframe=blue!50!black,
    colback=blue!5!white,
    coltitle=white
]
\label{box:reward}
\textbf{Input:} $feat_t$ (dim: 1536)

\begin{verbatim}
x = Linear(1536, 512) -> LayerNorm -> SiLU
x = Linear(512, 512) -> LayerNorm -> SiLU
x = Linear(512, 512) -> LayerNorm -> SiLU
x = Linear(512, 512) -> LayerNorm -> SiLU # Extra depth

logits = Linear(512, 255)
reward_dist = Softmax(logits)
\end{verbatim}
\end{tcolorbox}

\subsection{Continuation Head}
The Continuation Head predicts the probability $\hat{c}_t \in [0, 1]$ of the episode continuing at the next step (equivalent to $1 - \text{probability of termination}$). This allows the agent to ``hallucinate'' termination events, such as the robot falling over or exceeding joint limits, during latent imagination.

Unlike the Reward Head, the Continuation Head is relatively shallow (2 layers). We found that the termination boundary (alive vs. dead) is generally sharper and easier to learn than the complex reward landscape, requiring less network depth.

\begin{tcolorbox}[title={\textbf{Continuation Predictor}},
    colframe=blue!50!black,
    colback=blue!5!white,
    coltitle=white
]
\label{box:cont}
\textbf{Input:} $feat_t = [h_t, z_t]$ (dim: 1536)

\begin{verbatim}
# Layer 1
x = Linear(1536, 512) -> LayerNorm -> SiLU

# Layer 2
x = Linear(512, 512) -> LayerNorm -> SiLU

# Binary Output (Bernoulli Distribution)
logit = Linear(512, 1)
p_alive = Sigmoid(logit)
\end{verbatim}
\end{tcolorbox}

\clearpage
\section{Morphological Diversity Analysis}\label{app:morphology}
To characterize how each robot relates to the rest of the cohort on individual physical dimensions, we compute \textit{per-feature leave-one-out (LOO) $z$-scores}. We deliberately avoid collapsing this multi-dimensional physical heterogeneity into a single scalar distance metric. Different features (limb lengths, log-masses, torque densities, stance aspect ratios, discrete topology flags) affect locomotion dynamics non-uniformly, and any scalar summary, whether cosine similarity or Euclidean distance in normalized space, implicitly assumes a feature isotropy that does not hold in physical reality. Instead, we characterize each held-out robot by its full per-feature deviation profile, which retains the directional and magnitude information needed to interpret the zero-shot transfer results. For each robot $r$ and continuous feature $f$, the LOO $z$-score is:
\begin{equation}
    z_{r,f} = \frac{f_r - \hat{\mu}_{-r,f}}{\hat{\sigma}_{-r,f}}
    \label{eq:loo_zscore}
\end{equation}
where $\hat{\mu}_{-r,f}$ and $\hat{\sigma}_{-r,f}$ are the mean and standard deviation of feature $f$ computed over the remaining seven robots. Crucially, this seven-robot reference set identically matches the training cohort used when evaluating the held-out robot $r$ in our zero-shot experiments (\autoref{sec:generalization}). A large $|z|$ on a given feature indicates that the robot deviates substantially from what the rest of the cohort represents on that physical dimension. The discrete topology flag $k_\text{cfg}$ is excluded from this analysis as its binary nature makes a Gaussian deviation measure inappropriate.

\begin{figure}[ht]
    \centering
    \includegraphics[width=\linewidth]{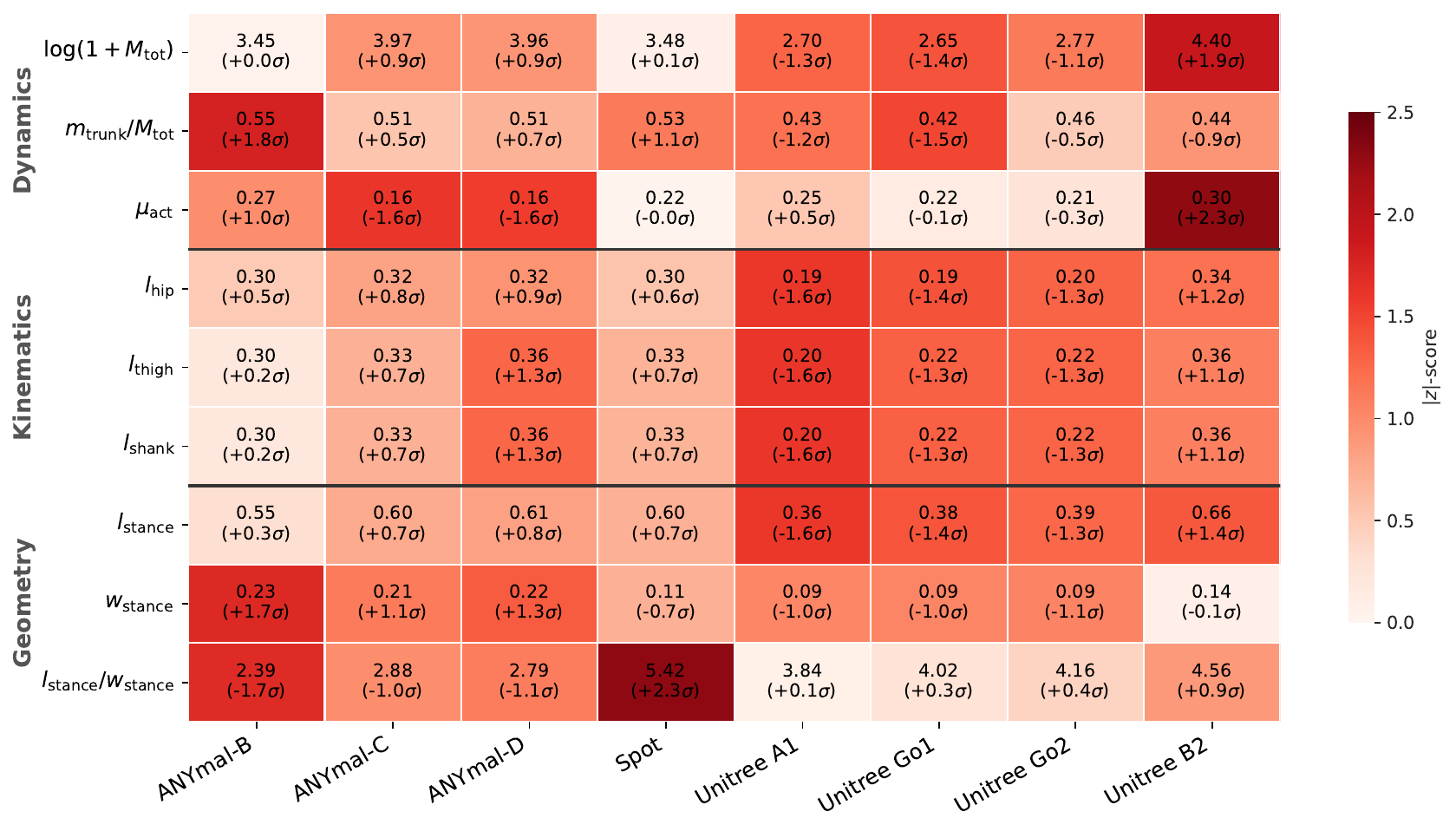}
    \caption{Per-feature LOO $z$-scores for the robot cohort. Each cell reports the raw physical value (top) and its signed deviation from the LOO cohort distribution in standard-deviation units (bottom). Color intensity encodes $|z|$, saturating at $2.5\sigma$. Features are grouped by category (Dynamics, Kinematics, Geometry). The discrete topology flag $k_\text{cfg}$ is excluded.}
    \label{fig:per_feature_ood}
\end{figure}

\autoref{fig:per_feature_ood} visualizes the signed $z$-scores. Each cell shows the raw value and its deviation in standard-deviation units. The color encodes $|z|$, saturating at $2.5\sigma$. \autoref{tab:loo_summary} summarizes the deviation statistics per robot.

\begin{table}[ht]
\centering
\footnotesize
\caption{Per-robot LOO $z$-score summary over 9 continuous features.}
\label{tab:loo_summary}
\begin{tabular}{@{}lcccc@{}}
\toprule
\textbf{Robot} & \textbf{Mean $|z|$} & \textbf{Max $|z|$} & \textbf{Worst feature}
               & \textbf{\# $|z|>1.5$} \\
\midrule
ANYmal-B   & 0.83 & 1.79 & $m_\text{trunk}/M_\text{tot}$ & 3/9 \\
ANYmal-C   & 0.90 & 1.61 & $\mu_\text{act}$            & 1/9 \\
ANYmal-D   & 1.10 & 1.56 & $\mu_\text{act}$            & 1/9 \\
Spot       & 0.77 & 2.30 & $l_\text{stance}/w_\text{stance}$ & 1/9 \\
Unitree A1 & 1.17 & 1.61 & $l_\text{thigh}$              & 4/9 \\
Unitree Go1& 1.09 & 1.49 & $m_\text{trunk}/M_\text{tot}$ & 0/9 \\
Unitree Go2& 0.94 & 1.30 & $l_\text{hip}$                & 0/9 \\
Unitree B2 & 1.20 & 2.29 & $\mu_\text{act}$            & 2/9 \\
\bottomrule
\end{tabular}
\end{table}

Three patterns are relevant to the zero-shot generalization results in \autoref{sec:generalization}.

\textbf{Unitree Go1.} No feature exceeds $1.5\sigma$ (max $|z|=1.49$ on trunk mass ratio). The Go1 is the most uniformly in-distribution robot in the cohort on all physical axes simultaneously. This directly explains its strong zero-shot performance.

\textbf{ANYmal-D.} One feature exceeds $1.5\sigma$ (torque capacity at $z=-1.6$), with all others below $1.3\sigma$. Like the Go1, the ANYmal-D sits comfortably within the support of the training distribution across the full feature vector, explaining its successful zero-shot transfer.

\textbf{Unitree B2.} The B2 shows the most broadly elevated deviation profile. Torque capacity reaches $z=+2.29$ (the only feature crossing $2\sigma$ in the cohort aside from Spot's aspect ratio), log-mass reaches $z=+1.92$, stance length $z=+1.37$, and hip offset $z=+1.19$. Critically, all elevated deviations push in the same direction, larger, heavier, and stronger, so no training robot simultaneously combines the B2's inertial regime, actuator capacity, and limb geometry. This is a \textit{combinatorial} gap. Individually each deviation is bounded, but jointly the configuration is unrepresented, which is why a scalar distance metric is insufficient to characterize B2's difficulty. A robot could match B2's scalar distance by deviating on only one or two features and would pose a qualitatively easier problem. That \ourMethod transfers to B2 zero-shot, while a model-free controller given the identical $\mu$ collapses (\autoref{sec:generalization}), indicates the learned dynamics combine mild extrapolation on mass and torque capacity with interpolation along the remaining physical axes, rather than relying on a single nearest neighbor.

\textbf{Boston Dynamics Spot.} Spot's deviation is narrow and concentrated, with aspect ratio reaching $z=+2.30$ (driven by its uniquely narrow stance, $l/w=5.42$), while all other features remain below $1.13\sigma$. As a \textit{training} robot, this one-dimensional geometric outlier status is handled well, since the WM only needs to adapt along a single physical axis to accommodate Spot's dynamics.

\clearpage
\section{Ablation Study}\label{app:ablation}

In this section, we rigorously evaluate the individual contributions of the proposed components in the \ourMethod framework. To validate our architectural design choices, specifically the Physical Morphology Encoder (PME), the Adaptive Reward Normalization (ARN), and the specific injection points of the morphology vector, we trained distinct variations of our model on the full heterogeneous robot cohort.

We compare the full \ourMethod framework against the following configurations to isolate the impact of (1) explicit morphology information, (2) reward balancing, and (3) the location of conditioning within the WM.

\begin{enumerate}
    \item \textbf{\ourMethod (Full Method).} The proposed framework as described in \autoref{sec:qwm}. The model utilizes the PME to generate $\mu$, applies ARN to normalize learning signals, and conditions both the dual-pathway encoder and the RSSM dynamics on $\mu$.

    \item \textbf{w/o PME (Implicit Identification).} We remove the explicit morphology conditioning entirely. The model receives no static physical parameters ($\mu$ is removed). The agent must implicitly infer the robot's morphology solely from the history of proprioceptive observations ($o_t$) stored in the recurrent state $h_t$. This effectively tests the limitations of implicit system identification. Note that the model still has the ARN module, making it different from the DreamerV3 backbone.

    \item \textbf{w/o ARN (Raw Rewards).} We remove the Adaptive Reward Normalization. The WM and policy are trained on the raw, unnormalized reward signals provided by the heterogeneous environments. This variant tests the hypothesis that varying reward scales across robots cause optimization instability or mode collapse.

    \item \textbf{w/o RSSM Conditioning.} We retain the PME and inject $\mu$ into the encoder, but we remove the explicit injection of $\mu$ into the recurrent model (\autoref{eq:rnn}). The transition dynamics become $h_t = f_\phi(h_{t-1}, z_{t-1}, a_{t-1})$. This tests whether the recurrent state $h_t$ can successfully maintain static morphological information over time without explicit re-injection at every step.

    \item \textbf{w/o Encoder Conditioning.} We retain the PME and inject $\mu$ into the recurrent model, but remove the static tower from the encoder. The posterior $z_t$ is inferred solely from dynamic observations $o_t$. This tests the necessity of grounding the latent representation $z_t$ with static physical context.

    \item \textbf{w/o ARN \& RSSM Conditioning.} A compound ablation removing both reward normalization and the explicit dynamics conditioning. This tests if reward scaling issues can be mitigated purely by the encoder's awareness of the morphology.

    \item \textbf{w/o ARN \& Encoder Conditioning.} A compound ablation removing both reward normalization and the encoder's access to morphology. This tests if the recurrent dynamics alone can handle both the unnormalized signal and the morphology retention.

    \item \textbf{w/o PME \& ARN (DreamerV3 Baseline).} We remove all proposed contributions. The model is a standard DreamerV3 \citep{hafner2025mastering} trained on the heterogeneous batch. It relies on raw rewards and implicit system identification. This serves as the lower-bound baseline to demonstrate the total value added by the \ourMethod framework.
\end{enumerate}

\begin{figure}[ht]
    \centering
    \includegraphics[width=\linewidth]{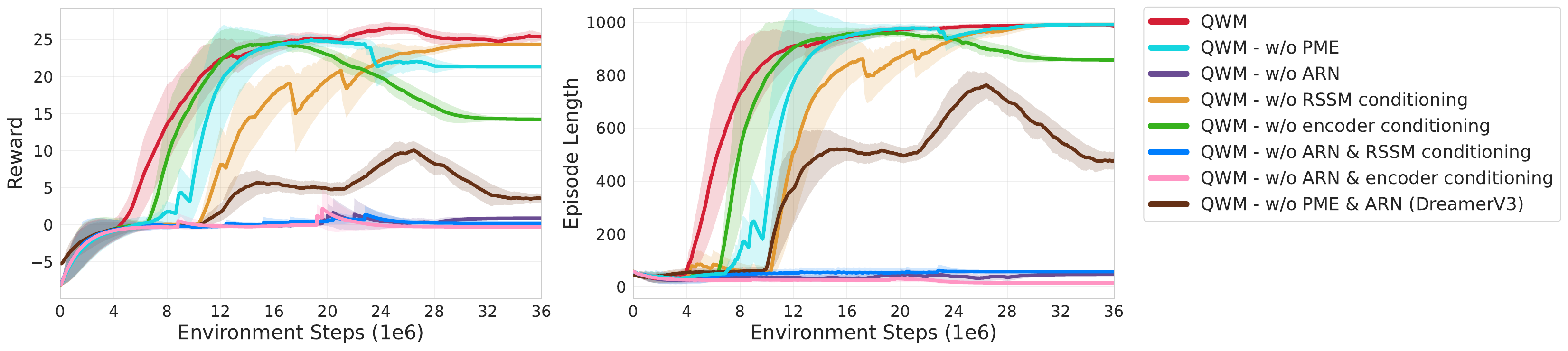}
    \caption{Ablation study on the heterogeneous cohort. We compare \ourMethod against architectural ablations on morphology encoding (PME), reward normalization (ARN), and conditioning locations. Shaded regions denote standard deviation across 5 seeds. \textbf{Key findings.} (1) \textcolor[HTML]{d41f35}{\ourMethod} achieves the fastest convergence and highest stable asymptotic performance. (2) \textcolor[HTML]{2ca02c}{Removing encoder conditioning} allows initial learning but leads to a catastrophic late-stage collapse. (3) \textcolor[HTML]{05a8b0}{Removing explicit morphology (PME)} maximizes episode length but plateaus at a suboptimal reward, showing explicit morphology is needed for refined gaits. (4) \textcolor[HTML]{6121b8}{Removing ARN} prevents learning, leaving a flatline near zero reward.}
    \label{fig:ablation}
\end{figure}

\autoref{fig:ablation} demonstrates a thorough comparison of learning performance across all ablated configurations. The results provide empirical evidence for the necessity of each component in the \ourMethod framework.

First, the most striking observation is the absolute necessity of Adaptive Reward Normalization (ARN) when training across a heterogeneous fleet. All configurations lacking ARN, specifically \textbf{w/o ARN}, \textbf{w/o ARN \& RSSM conditioning}, and \textbf{w/o ARN \& encoder conditioning}, fail entirely to learn. These variants remain flat near zero reward and minimal episode lengths throughout the 36 million environment steps. This confirms the hypothesis that unnormalized, disparate reward scales across different physical embodiments cause severe optimization instability, effectively preventing the world model from establishing a useful foundation.

Second, isolating the injection points for morphological conditioning reveals distinct failure modes when either pathway is omitted. Removing explicit morphology from the static tower of the encoder (\textbf{w/o encoder conditioning}) allows the model to learn initially, reaching competitive performance by 14 million steps. However, it subsequently suffers a catastrophic collapse, losing significant reward and episode length. This indicates that without grounding the latent posterior in static physical context, the representations become fundamentally unstable over extended training horizons. Conversely, removing explicit conditioning from the recurrent dynamics (\textbf{w/o RSSM conditioning}) heavily degrades sample efficiency and introduces substantial variance. While the model eventually achieves maximum episode lengths, its asymptotic reward remains sub-optimal, demonstrating that the recurrent state struggles to maintain the necessary morphological context over time without step-wise re-injection.

Third, we evaluate the limitations of purely implicit system identification. The \textbf{w/o PME} variant, which attempts to infer morphology solely from proprioceptive history without explicit physical parameters, successfully learns to maximize episode length but plateaus at a noticeably lower asymptotic reward compared to the full method. This performance gap highlights that while implicit system identification can achieve basic locomotion, explicit physical morphology is required to discover and stabilize highly refined, optimal gaits across diverse robots and to improve sample efficiency.

Finally, the standard baseline (\textbf{w/o PME \& ARN}, structurally equivalent to DreamerV3) performs poorly on the joint heterogeneous distribution. It peaks at a reward of roughly 10 and fails to achieve stable, full-length episodes before performance begins to degrade. In stark contrast, the full \textbf{\ourMethod} framework consistently demonstrates the fastest sample efficiency, lowest variance, and highest asymptotic performance, validating that the integrated combination of explicit morphology encoding, dual-pathway conditioning, and reward normalization is critical for successful heterogeneous control.

\clearpage
\section{Real-World Deployment Details}
\label{app:realworld}

We provide a quantitative assessment of the real-world deployment described in \autoref{sec:realworld}. Our goal is to measure how closely the zero-shot \ourMethod policy tracks velocity commands on physical hardware compared to a specialist PPO controller trained exclusively on each target platform, which serves as the empirical upper bound.

\paragraph{Protocol.} We conduct 10 trials per platform, each lasting 60 seconds, with random velocity commands sampled uniformly from the training command ranges (\autoref{tab:commands}). Trials are conducted on flat indoor ground. The specialist PPO baseline is evaluated under identical conditions. No falls or manual resets occurred in any trial for either method. We report mean absolute tracking error for linear velocity ($e_{xy}$, m/s) and angular velocity ($e_{yaw}$, rad/s), averaged across all timesteps and trials.

\paragraph{Discussion.} \ourMethod achieves tracking errors within $\sim$10\% of the specialist baseline on both platforms. The gap stays bounded across both metrics and both robots, suggesting a systematic but bounded performance cost from zero-shot transfer rather than catastrophic failure. The complete absence of falls across all 20 trials confirms that the explicit morphology conditioning encodes sufficient physical grounding to maintain stability on unseen hardware in the presence of real-world unmodeled dynamics such as actuator backlash and ground friction variability. The slightly larger gap on the Unitree Go1 ($e_{xy}$: $0.34$ vs. $0.31$) compared to ANYmal-D ($e_{xy}$: $0.30$ vs. $0.28$) is consistent with the Go1's higher-frequency dynamics, which place greater demands on precise timing of contact events that the zero-shot policy has not been tuned for.

\clearpage
\section{Latent State Disentanglement Analysis}
\label{app:latent_viz}

A core architectural claim of \ourMethod is that explicitly conditioning $h_t$ on $\mu$ at every recurrent step disentangles morphological identity from dynamic state. The recurrent state $h_t$ should carry the static physical context, while $z_t \sim q_\phi(z_t \mid h_t, e_t)$, freed from re-inferring morphology from observations, should encode only the current dynamic state. This appendix provides the complete analysis testing this claim directly, with dimensionality reduction visualizations and quantitative probing across morphological and dynamic variables.

\subsection{Data Collection and Protocol}

To collect a representative sample of the learned latent space, we roll out the trained \ourMethod policy on all eight robots in our cohort under a uniform random velocity-command distribution. We collect latent pairs $(h_t, z_t)$ from $128$ independent trajectories of $32$ steps each, yielding $4096$ latent state pairs per robot and $32768$ points in total.
At each step we record the full concatenated latent vector $[h_t, z_t] \in \mathbb{R}^{3072}$ alongside scalar annotations, namely robot identity (morphology label), forward speed $v_x$, linear speed $\|v\|$, angular speed $\|\omega\|$, and mean joint activity $\text{mean}|q|$ as a proxy for gait phase.
These annotations are used purely for post-hoc coloring and play no role in the model.

\subsection{Morphology Disentanglement of \texorpdfstring{$h_t$}{h\_t} and \texorpdfstring{$z_t$}{z\_t}}
\label{app:latent_viz:morphology}

\begin{figure}[ht]
    \centering
    \includegraphics[width=0.6\linewidth]{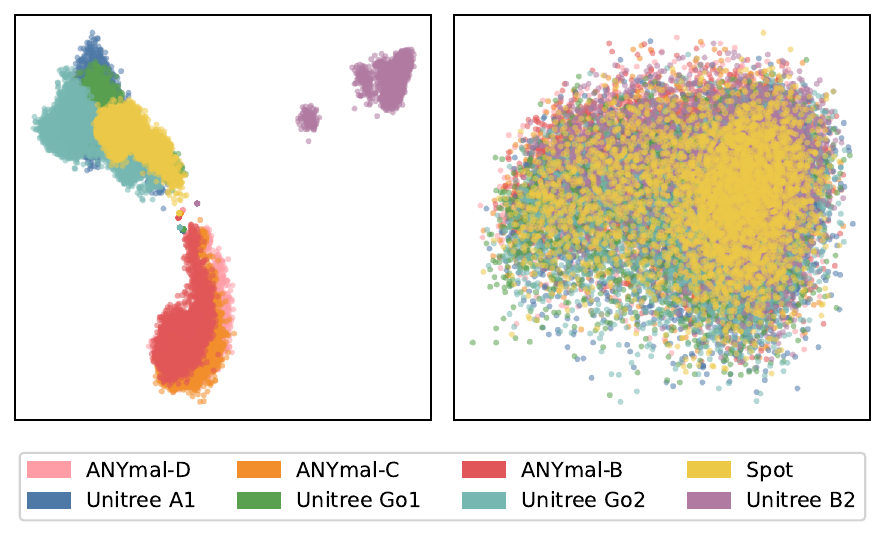}
    \caption{
        Latent disentanglement. PCA of $h_t$ (left) and $z_t$ (right), both colored by robot identity. The $\mu$-conditioned recurrent state organizes into morphology-specific clusters. The stochastic state collapses into a single morphology-agnostic cloud. By morphology the silhouette score of $h_t$ greatly exceeds that of $z_t$, quantifying the asymmetry the architecture is designed to produce.
    }
    \label{fig:disentangle_main}
\end{figure}

Applying PCA to each component separately, the contrast in \autoref{fig:disentangle_main} is direct evidence for the intended functional decomposition. The $\mu$-conditioned recurrent state $h_t$ partitions cleanly into morphology-specific clusters, with each robot occupying a distinct region of the projection. The stochastic state $z_t$, by contrast, collapses into a single intermixed cloud with no visible morphological structure, yet retains smooth velocity gradients under the same projection (\aref{app:latent_viz:probing}), confirming that dynamic information is preserved.

A probing analysis quantifies the asymmetry. The silhouette score of $z_t$ under morphology labels is $0.033$ with only $9.8\%$ between-class variance (detailed further in \aref{app:latent_viz:metrics}), indicating that morphological identity is not redundantly re-encoded in $z_t$ when $\mu$ is provided to $h_t$. This decomposition is not imposed by any auxiliary loss. It arises naturally from the explicit conditioning, and it underpins \ourMethod's ability to control unseen robots by swapping $\mu$ at deployment. Crucially, it also explains the robustness demonstrated in \aref{app:mu_perturbation}, because $z_t$ encodes dynamic state independently of morphological identity, so an inaccurate $\mu$ that mis-steers $h_t$ can be partially corrected online through $z_t$'s conditioning on real observations $o_t$, providing a self-correcting mechanism absent from any purely parameter-conditioned system.

\subsection{PCA of the Full Latent State}
\label{app:latent_viz:pca}

\begin{figure*}[ht]
    \centering
    \includegraphics[width=\linewidth]{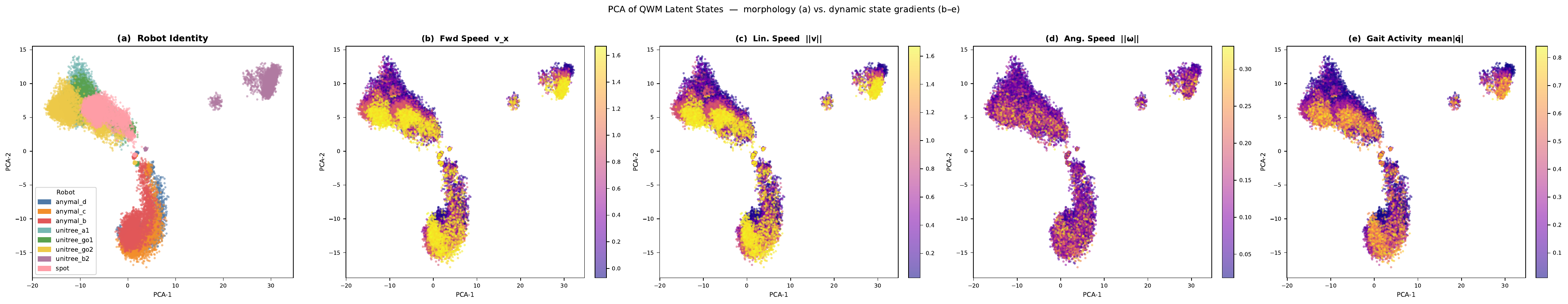}
    \caption{
        \textbf{PCA of \ourMethod latent states} $[h_t, z_t]$, morphology (a) vs. dynamic state gradients (b-e).
        Each point is one latent observation, and $32768$ points are shown across all eight robots.
        (a) Coloring by robot identity reveals that the latent space organizes itself into morphology-specific clusters, even though robot identity is \textit{never} provided as a supervised signal.
        (b-e) The same projection colored by four continuous dynamic variables shows smooth, consistent gradients \textit{within} each morphology cluster, demonstrating that dynamic execution context is encoded orthogonally to morphological identity.
        PCA explained variance for the first two components is $7.9\%$, reflecting the high-dimensional nature of the full latent space.
    }
    \label{fig:pca_combined}
\end{figure*}

\autoref{fig:pca_combined} presents PCA of the full $[h_t, z_t]$ latent vector projected onto its first two principal components.
Panel (a) colors each point by robot identity.
The three ANYmal variants (B, C, D) form an overlapping super-cluster, reflecting their high pairwise $\mu$ similarity in the morphology feature space.
Similarly, the Unitree series (A1, Go1, Go2) and Spot occupy a contiguous band, while Unitree B2 occupies a well-isolated region in the upper right of the PCA plane.
These groupings emerge unsupervised, providing strong evidence that the latent dynamics model has internalized the physical kinematic families present in the training cohort. Notably, this isolated cluster remains controllable by the policy (\autoref{tab:zero_shot_results}), indicating that the learned latent geometry retains predictive validity even in low-density regions of $\mu$-space.

Panels (b-e) overlay continuous dynamic annotations on the same projection.
Within each morphology cluster, there are smooth, spatially coherent gradients along all four dynamic variables.
High forward speed $v_x$ (yellow) consistently occupies specific sub-regions of each cluster, transitioning smoothly to low-speed states (purple) within the same cluster boundary.
The same pattern holds for linear speed $\|v\|$, angular speed $\|\omega\|$, and gait activity $\text{mean}|q|$.
This indicates that the learned latent geometry encodes dynamic state as a \textit{continuous manifold within} each morphology's cluster, rather than conflating the two sources of variation into an entangled representation.

The relatively low explained variance of the first two PCA components ($7.9\%$) is expected, since the latent space is $3072$-dimensional and captures rich, high-frequency dynamics.
The visible structure in PCA-2 is therefore a conservative lower bound on the actual degree of organization present.

\subsection{t-SNE of the Full Latent State}
\label{app:latent_viz:tsne}

\begin{figure*}[ht]
    \centering
    \includegraphics[width=\linewidth]{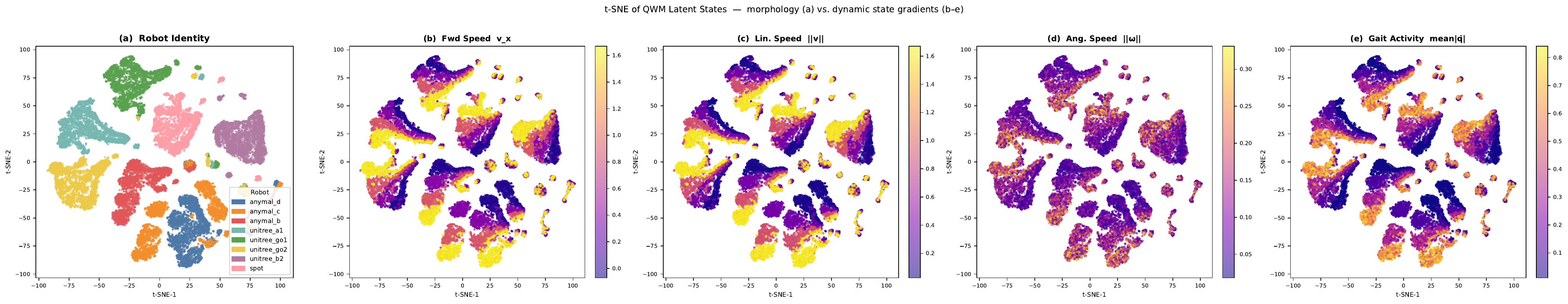}
    \caption{
        \textbf{t-SNE of \ourMethod latent states} $[h_t, z_t]$, morphology (a) vs. dynamic state gradients (b–e).
        t-SNE (perplexity $= 40$, $1000$ iterations, KL divergence $= 1.73$) is applied to the same $32768$-point dataset as \autoref{fig:pca_combined}.
        (a) Each robot occupies a spatially distinct, often multi-lobed cluster, and the multi-lobe structure within a single robot reflects different locomotion regimes (e.g., slow walk vs. fast trot).
        (b–e) Dynamic variables exhibit strong intra-cluster gradients, confirming that within each morphology's region the model encodes a continuous dynamic state manifold.
    }
    \label{fig:tsne_combined}
\end{figure*}

\autoref{fig:tsne_combined} presents a t-SNE embedding (perplexity $= 40$, $1000$ iterations, final KL divergence $= 1.73$) of the same dataset.
t-SNE preserves local neighborhood structure and thus reveals finer-grained organization than PCA.

Panel (a) shows that every robot occupies one or more spatially isolated regions with near-zero overlap between morphology classes.
Several robots exhibit a characteristic multi-lobe structure, for instance the ANYmal variants each show a main cluster flanked by smaller satellite regions.
We attribute these sub-clusters to distinct locomotion regimes captured by the policy (e.g., stationary standing, slow-speed trotting, high-speed locomotion), consistent with the dynamic gradients visible in panels (b-e).

The Unitree A1, Go1, and Go2 clusters, while distinct in panel (a), are spatially adjacent in the t-SNE plane, mirroring their high z-score similarity in the morphology feature space.
Analogously, the ANYmal-B, C, and D clusters are neighboring, consistent with z-score similarities.
This spatial proximity in the learned latent space directly explains \ourMethod's success at zero-shot interpolation (\autoref{sec:generalization}). The target robot's morphology vector places it in a high-density region of the training distribution, allowing the WM to synthesize a plausible dynamic model by interpolating between its nearest neighbors in latent space.

Panels (b-e) confirm the same pattern observed in PCA, with velocity gradients smooth and spatially coherent within each morphological cluster.
Importantly, no systematic leakage of dynamic information \textit{across} morphology boundaries is observed. Regions of high $v_x$ for one robot type do not merge with high-$v_x$ regions of a different robot type, indicating that dynamic and morphological degrees of freedom are not collapsed into the same latent directions.

\subsection{Probing Analysis of \texorpdfstring{$h_t$}{h\_t} and \texorpdfstring{$z_t$}{z\_t}}
\label{app:latent_viz:probing}

\begin{figure*}[ht]
    \centering
    \includegraphics[width=0.6\linewidth]{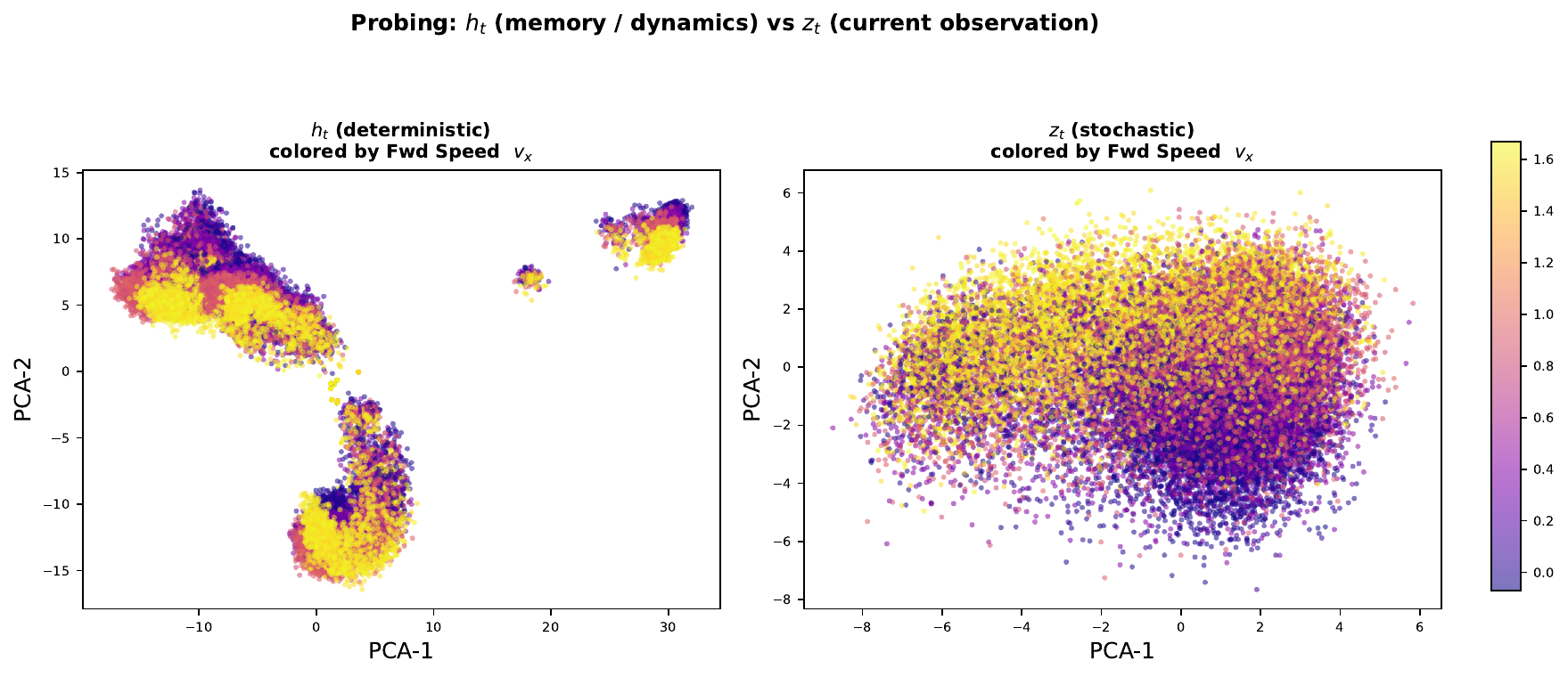}
    \caption{
        Dynamic-state probing of $h_t$ vs. $z_t$.
        PCA projections of each latent component, colored by forward speed $v_x$. The morphology-colored counterpart of this figure appears earlier in this section (\autoref{fig:disentangle_main}), and here we show the same projections under a dynamic-state coloring to complete the probing picture. Left ($h_t$): Velocity gradients are visible \textit{within} each morphology-specific region, indicating that $h_t$ tracks the evolving dynamic trajectory of each robot in its morphology-aligned subspace. Right ($z_t$): The cloud that is unstructured by morphology (see \autoref{fig:disentangle_main}), yet a smooth $v_x$ gradient persists across the full projection, confirming that $z_t$ retains the instantaneous dynamic state of the robot even as it sheds morphological identity.
    }
    \label{fig:h_vs_z_probing}
\end{figure*}

\aref{app:latent_viz:morphology} introduced the central contrast. Under PCA, $h_t$ partitions into morphology-specific clusters while $z_t$ collapses into a single morphology-agnostic cloud. That analysis established \textit{which component carries morphological identity}. Here we complete the probing picture by overlaying \textit{dynamic-state} information on the same projections, asking the complementary question of whether each component retains the current dynamic state or loses it when morphology is removed.

\textbf{The deterministic recurrent state $h_t$.} The left panel of \autoref{fig:h_vs_z_probing} shows that $h_t$ encodes \textit{both} morphological identity (visible as cluster structure in \autoref{fig:disentangle_main}) and dynamic state (visible as the $v_x$ gradient within each cluster). This dual structure is the intended behavior of the architecture. The recurrent state $h_t = f_\phi(h_{t-1}, z_{t-1}, a_{t-1}, \mu)$ is conditioned on $\mu$ at every step, continuously steering the hidden state toward a morphology-aligned regime, while the GRU recurrence integrates the action and prior stochastic state to track the dynamic trajectory within that regime. The result is a \textit{physics-contextualized memory}. $h_t$ knows which robot it is controlling and follows that robot's evolving state through its morphology-specific region of latent space.

\textbf{The stochastic observation state $z_t$.} The right panel presents the critical evidence. Although $z_t$ has shed all morphological structure (\autoref{fig:disentangle_main}), a smooth, spatially coherent $v_x$ gradient remains visible across the projection, with high-velocity states (yellow) on one side of the cloud and low-velocity states (purple) on the other and continuous transitions between them. The dynamic information has not been lost, it has been \textit{decoupled} from morphological identity. This is what the architecture predicts. Because $\mu$ is explicitly provided to $h_t$, the stochastic encoder $z_t \sim q_\phi(z_t \mid h_t, e_t)$ does not need to re-infer physical identity from the observation $o_t$, since that context is already present in $h_t$. Consequently $z_t$ is free to encode only the residual, morphology-independent dynamic information in the current observation. The two views together (morphology in \autoref{fig:disentangle_main}, velocity here) form the complete probing argument. $h_t$ carries both signals, $z_t$ carries only the dynamic one.

\subsection{Quantitative Disentanglement Metrics}
\label{app:latent_viz:metrics}

To complement the visual evidence, we compute three quantitative metrics on the $z_t$ embeddings, the latent component that, according to our hypothesis, should be \textit{free} of morphological structure.

\begin{table}[ht]
\centering
\caption{Quantitative disentanglement metrics computed on the stochastic state $z_t$ across all $32{,}768$ collected points. Low silhouette score and high within-class variance fraction confirm that $z_t$ encodes negligible morphological information.}
\label{tab:disentanglement_metrics}
\begin{tabular}{lcc}
\toprule
\textbf{Metric} & \textbf{Value} & \textbf{Interpretation} \\
\midrule
Silhouette score (morphology) & $0.033$ & $\approx 0$ $\Rightarrow$ no cluster separation \\
Between-class variance fraction & $0.098$ & $9.8\%$ of variance is inter-morphology \\
Within-class variance fraction  & $0.901$ & $90.1\%$ of variance is intra-morphology \\
\bottomrule
\end{tabular}
\end{table}

\autoref{tab:disentanglement_metrics} reports these metrics for $z_t$.
The silhouette score of $0.033$ (range $[-1, 1]$, where $1$ indicates perfect separation and $0$ indicates random overlap) is near zero, indicating that morphology labels are essentially meaningless for partitioning the $z_t$ space.
The variance decomposition reinforces this, with only $9.8\%$ of total variance in $z_t$ falling between morphology classes while $90.1\%$ is within-class dynamic variance.
Taken together, these numbers confirm that the stochastic state $z_t$ carries overwhelmingly more information about the robot's instantaneous dynamic state than about its physical identity.

We note that computing the same silhouette score on $h_t$ yields a substantially higher value, consistent with the visible cluster separation in the top row of \autoref{fig:h_vs_z_probing}.
The asymmetry between $h_t$ and $z_t$ on this morphology-based metric is the quantitative signature of the disentanglement the architecture is designed to achieve.

\subsection{Summary}

Converging evidence from PCA (\aref{app:latent_viz:pca}), t-SNE (\aref{app:latent_viz:tsne}), probing (\aref{app:latent_viz:probing}), and quantitative metrics (\aref{app:latent_viz:metrics}) confirms the functional decomposition introduced in \aref{app:latent_viz:morphology}. The $\mu$-conditioned recurrent state $h_t$ encodes both morphological identity and dynamic trajectory context, while the stochastic state $z_t$ encodes purely dynamic state with negligible morphological structure (silhouette $= 0.033$, $9.8\%$ between-class variance). The spatial proximity of kinematically similar robots in the t-SNE plane mirrors their low pairwise distance in morphology feature space, directly explaining \ourMethod's interpolation capability. This decomposition arises naturally from the explicit $\mu$ conditioning, without any auxiliary disentanglement loss.

\clearpage
\section{Robustness to Perturbations}\label{app:perturbation}

The zero-shot generalization results (\autoref{sec:generalization}) show the model-based route succeeding where a model-free policy given the same $\mu$ does not. The underlying mechanism is a correction channel that conditioning the dynamics creates and conditioning a policy cannot. PME-PPO maps $\mu$ directly to actions, so any error in $\mu$, or any drift of the real robot away from its specification, propagates straight to control with no correction. \ourMethod instead exposes its policy to the latent state $(h_t, z_t)$, where $\mu$ sets the morphological prior through $h_t$ while the stochastic state $z_t \sim q_\phi(z_t \mid h_t, e_t)$ re-reads the true dynamic state from observations $o_t$ at every step. Because $z_t$ is empirically decoupled from morphological identity (\aref{app:latent_viz}), it can encode the residual between the specified and the realized robot, which a feedforward $\mu$-policy has no channel to represent.

We probe this mechanism along two orthogonal axes. \aref{app:mu_perturbation} holds the world nominal and corrupts $\mu$ itself, asking whether the latent channel absorbs errors in the specification. \aref{app:unmodeled} holds $\mu$ nominal and perturbs the true world, asking whether the latent channel absorbs physical effects that $\mu$ cannot describe. Together the two studies bracket both directions of spec-versus-reality divergence.

\subsection{Robustness to Morphological Parameter Uncertainty}\label{app:mu_perturbation}

We argue that the learned latent dynamics $(h_t, z_t)$ capture behavior that $\mu$ cannot encode. PME-PPO is our closest proxy for a purely spec-sheet-driven controller, since it receives the same $\mu$ but has no learned residual channel. If the latent-channel hypothesis holds, \ourMethod should tolerate minor $\mu$ inaccuracies such as unmodeled payloads or miscalibration, because the latent channel absorbs the residual. A pure lookup has no such fallback. We perturb $\mu$ at test time by applying multiplicative noise to each raw feature before normalization:
\begin{equation}
    f_i^{(\delta)} = f_i \cdot (1 + \epsilon_i), \quad \epsilon_i \sim
    \mathcal{U}(-\delta,\, \delta), \quad \delta \in \{0.05, 0.10, 0.20\}
    \label{eq:mu_perturb}
\end{equation}
We then re-apply training-cohort normalization, so the model receives a plausible but inaccurate $\mu$ rather than an out-of-range input. The discrete topology flag $k_\text{cfg}$ is held fixed, since flipping it changes the kinematic family entirely. We evaluate over 20 noise draws $\times$ 5 seeds $\times$ 128 envs on all three held-out robots.

\begin{wraptable}{r}{0.5\textwidth}
\centering
\scriptsize
\renewcommand{\arraystretch}{0.85}
\setlength{\tabcolsep}{2.5pt}
\caption{Robustness to $\mu$ perturbation (episode length under multiplicative noise on raw features). \ourMethod degrades gracefully. PME-PPO fails sharply, lacking a latent correction channel. Mean $\pm$ std over 20 draws $\times$ 5 seeds. $\delta=0$ reproduces \autoref{tab:zero_shot_results}.}
\label{tab:mu_perturbation}
\begin{tabular}{@{}llccc@{}}
\toprule
\textbf{Method} & $\delta$ & \textbf{ANYmal-D} & \textbf{Unitree Go1} &
\textbf{Unitree B2} \\
\midrule
\multirow{4}{*}{\textbf{PME-PPO}}
  & $0.00$ & $530 \pm 8.2$   & $602 \pm 14.9$  & $337 \pm 23.1$ \\
  & $0.05$ & $441 \pm 18.4$   & $498 \pm 22.7$   & $284 \pm 31.8$ \\
  & $0.10$ & $329 \pm 27.1$   & $387 \pm 31.5$   & $223 \pm 38.4$ \\
  & $0.20$ & $174 \pm 41.6$   & $208 \pm 47.2$   & $142 \pm 44.9$ \\
\midrule
\multirow{4}{*}{\textbf{\ourMethod}}
  & $0.00$ & $948.6 \pm 12.1$ & $974.4 \pm 6.2$ & $925.4 \pm 17.2$ \\
  & $0.05$ & $931.2 \pm 14.8$ & $962.1 \pm 9.4$  & $897.6 \pm 22.5$ \\
  & $0.10$ & $887.4 \pm 23.6$ & $921.7 \pm 18.1$ & $834.2 \pm 35.1$ \\
  & $0.20$ & $742.8 \pm 51.3$ & $801.5 \pm 43.7$ & $658.9 \pm 67.4$ \\
\bottomrule
\end{tabular}
\end{wraptable}

\autoref{tab:mu_perturbation} reveals an asymmetry growing with perturbation magnitude. At $\delta=0.05$, \ourMethod is essentially unaffected, dropping by at most $3\%$, within the range absorbed by domain randomization, while PME-PPO already loses $16-17\%$ of its lower baseline. At $\delta=0.1$, \ourMethod retains about $93\%$ against PME-PPO's $63\%$, and at $\delta=0.2$, about $80\%$ against $33\%$. \ourMethod never falls below the \textit{unperturbed} PME-PPO baseline on any robot, even when its own $\mu$ is corrupted by $20\%$. The mechanism distinguishes two architectures clearly. PME-PPO maps $\mu$ directly to actions, so parameter error propagates to control with no correction. \ourMethod instead exposes its policy to $(h_t, z_t)$, where $z_t$ is inferred from real observations $o_t$ at every step. Under a perturbed $\mu$, the recurrent state $h_t$ is mis-steered, but $z_t$ still reads the true dynamic state. The latent disentanglement in \aref{app:latent_viz} (silhouette $=0.033$, with $z_t$ encoding dynamic state independently of morphological identity) is the necessary precondition, because $z_t$ cannot compensate for an inaccurate $\mu$ unless it is decoupled from it. B2 sharpens this further. \ourMethod holds its episode length above 650 even at $\delta=0.2$, despite B2's unperturbed $\mu$ already lying outside the training cluster, whereas PME-PPO stays in the $140-280$ range across all $\delta$, below its operational floor regardless of $\mu$ quality. Together with the out-of-cluster B2 result (\autoref{sec:generalization}), where PME-PPO collapses while \ourMethod succeeds, and the disentanglement analysis (\aref{app:latent_viz}), this forms a three-part argument. First, $\mu$ alone is insufficient for out-of-cluster transfer. Second, $z_t$ carries dynamic information decoupled from $\mu$. Third, this decoupling absorbs $\mu$ inaccuracies at test time, a property structurally impossible for a spec-sheet lookup and exactly what a grey-box analytical-plus-residual approach would need, except that \ourMethod learns both channels jointly without a differentiable physics engine.

\subsection{Robustness to Unmodeled Dynamics}\label{app:unmodeled}
\aref{app:mu_perturbation} probed an \textit{inaccurate} $\mu$ on a nominal world. We now probe the complement, a \textit{correct} $\mu$ describing a robot whose true dynamics have departed from spec through effects $\mu$ cannot encode. This is the regime that motivates a learned dynamics model over an analytical engine, and we make the contrast explicit by adding a third baseline. \textbf{Analytical-$\mu$} is a rigid-body simulator used as an open-loop \textit{predictor} (not a controller). Given an initial state and an action sequence, it forward-simulates the resulting state trajectory from physics. We configure it deliberately to \textit{steelman} the spec-sheet case. It is instantiated with the robot's true \textit{nominal} parameters, including physical quantities that $\mu$ omits entirely, such as surface friction and joint damping. It therefore receives strictly more information than a real spec sheet provides, and is near-exact when the robot matches its nominal description. The one thing it is never told is the test-time perturbation. When we attenuate torque or add payload to the \textit{true} world, Analytical-$\mu$ continues to simulate the unperturbed nominal robot. It has no channel to reconcile its forecast against the robot's observed behavior.

Critically, Analytical-$\mu$ is \textit{not} online SysID, which would re-estimate parameters from observed transitions and, given enough data, correct for the perturbation. It is the degenerate case in which those parameters stay frozen at the nominal specification, isolating the predictive value of parameter conditioning alone. It is the analytical counterpart of PME-PPO, the latter being a model-free policy conditioned on $\mu$ and the former an analytical predictor conditioned on $\mu$. Both lack the learned latent residual $(h_t, z_t)$ that lets \ourMethod reconcile specified against realized dynamics, the correction a SysID module would otherwise provide, here amortized into a single frozen forward model rather than a per-step estimation loop.

\textbf{Open-loop prediction protocol.} For each perturbed rollout we run \ourMethod's policy on the true perturbed robot, record the executed action sequence and the ground-truth state trajectory, then ask each predictor to reproduce that trajectory from a short context window ($T=5$) over a 45-step horizon, feeding only the recorded actions. \ourMethod predicts via its transition model conditioned on nominal $\mu$. Analytical-$\mu$ predicts via rigid-body simulation of the nominal robot. We report NMSE against the perturbed ground truth, normalized by each robot's natural motion variance, in \autoref{tab:unmodeled_nmse}. All perturbations are applied at test time to frozen policies within the same environment used for training. Analytical-$\mu$ is a separate instance of the same PhysX engine configured as the nominal robot. Implementation details are given in \aref{app:unmodeled_protocol}.

\textbf{Torque attenuation.} We scale the realized joint torque, $\tau^{\text{eff}} = (1-\eta)\,\tau$, with $\eta \in \{0.1, 0.2\}$ applied uniformly across all 12 joints. This is the controlled single-parameter version of motor wear, which is typically asymmetric, and we leave per-joint degradation to future work. Crucially, $\mu$ stays nominal. The torque capacity feature $\mu_{\text{act}}$ still advertises the full robot description limits, exactly as a real deployment would, and the effect is absent from both $\mu$ and the training-time randomization (\aref{app:env_details}), making it genuinely unmodeled for every method.

\textbf{Payload.} We add a trunk payload equal to a fraction $p \in \{0.1, 0.2\}$ of the nominal mass, with a corresponding center-of-mass offset, again holding $\mu$ nominal. This is the mirror image of \aref{app:mu_perturbation}, where the dynamics were correct and $\mu$ was corrupted. Here $\mu$ is correct for the spec but the physical robot has departed from it. Together the two studies bracket both directions of spec-versus-reality divergence. We evaluate all three held-out robots over 20 draws $\times$ 5 seeds $\times$ 128 environments, reporting open-loop NMSE (\autoref{tab:unmodeled_nmse}) and closed-loop episode length (\autoref{tab:unmodeled_control}).

\textbf{The crossover isolates the learned dynamics.} \autoref{tab:unmodeled_nmse} shows the open-loop prediction comparison. At nominal conditions ($\eta=0$, $p=0$), Analytical-$\mu$ is near exact and outperforms \ourMethod, as expected when the true parameters are known and held fixed. The decisive observation is the crossover. As soon as the true dynamics depart from $\mu$, even mildly ($\eta=0.1$ or $p=0.1$), the analytical error climbs steeply because it keeps forward-simulating a robot that no longer exists, while \ourMethod stays low, with the same pattern at the stronger level. No fixed parameter set can track a world that has moved away from it, so the learned dynamics act as a residual-absorption mechanism that a static prior cannot supply.

\begin{wraptable}{r}{0.5\textwidth}
\centering
\scriptsize
\renewcommand{\arraystretch}{0.85}
\setlength{\tabcolsep}{2.5pt}
\caption{Open-loop NMSE under unmodeled dynamics, averaged over the three held-out robots (45-step horizon, normalized by motion variance). Analytical-$\mu$ is near exact at nominal but diverges once the true world departs from spec. \ourMethod stays bounded because $z_t$ reads the realized dynamics from observations.}
\label{tab:unmodeled_nmse}
\begin{tabular}{@{}llcc@{}}
\toprule
\textbf{Perturb.} & \textbf{Level} & \textbf{Analytical-$\mu$} & \textbf{\ourMethod} \\
\midrule
\multirow{3}{*}{\shortstack[l]{Torque\\ ($\eta$)}}
  & $0.0$ & $0.03 \pm 0.01$ & $0.11 \pm 0.03$ \\
  & $0.1$ & $0.34 \pm 0.06$ & $0.16 \pm 0.04$ \\
  & $0.2$ & $0.72 \pm 0.11$ & $0.24 \pm 0.05$ \\
\midrule
\multirow{3}{*}{\shortstack[l]{Payload\\ ($p$)}}
  & $0.0$ & $0.03 \pm 0.01$ & $0.11 \pm 0.03$ \\
  & $0.1$ & $0.28 \pm 0.05$ & $0.15 \pm 0.03$ \\
  & $0.2$ & $0.61 \pm 0.09$ & $0.21 \pm 0.05$ \\
\bottomrule
\end{tabular}
\end{wraptable}

\textbf{Closed-loop control mirrors this.} \autoref{tab:unmodeled_control} shows the same asymmetry in deployment. Under the strongest attenuation ($\eta=0.2$), \ourMethod retains 75 to 80\% of its nominal episode length while PME-PPO falls to 44 to 47\%, and at $p=0.2$ it retains 79 to 84\% against PME-PPO's 50 to 58\%. PME-PPO's degradation is partly expected, since a reactive model-free policy has no internal model against which to reconcile observed behavior. The informative signal is therefore the margin and shape of the gap, not its existence. \ourMethod exposes its policy to $(h_t, z_t)$, where $z_t \sim q_\phi(z_t \mid h_t, e_t)$ continuously re-estimates the true dynamic state from observations and fuses it with the learned model, absorbing the spec-versus-reality discrepancy online rather than propagating it to control.

\begin{wraptable}{r}{0.58\textwidth}
\centering
\scriptsize
\renewcommand{\arraystretch}{0.95}
\setlength{\tabcolsep}{3pt}
\caption{Robustness to unmodeled dynamics. Closed-loop episode length under torque attenuation and payload, with nominal $\mu$. \ourMethod degrades gracefully, since $z_t$ reads true dynamics from $o_t$. PME-PPO, lacking a learned model, fails sharply. Mean $\pm$ std over 20 draws $\times$ 5 seeds. The $0.0$ rows reproduce \autoref{tab:zero_shot_results}.}
\label{tab:unmodeled_control}
\begin{tabular}{@{}llccc@{}}
\toprule
\textbf{Method} & \textbf{Level} & \textbf{ANYmal-D} & \textbf{Unitree Go1} & \textbf{Unitree B2} \\
\midrule
\multicolumn{5}{l}{\textit{Torque Attenuation} ($\eta$)} \\
\multirow{3}{*}{PME-PPO}
  & $0.0$ & $530 \pm 8.2$    & $602 \pm 14.9$   & $337 \pm 23.1$ \\
  & $0.1$ & $388 \pm 21.4$   & $447 \pm 26.8$   & $233 \pm 31.6$ \\
  & $0.2$ & $244 \pm 33.7$   & $281 \pm 37.2$   & $148 \pm 36.9$ \\
\midrule
\multirow{3}{*}{\ourMethod}
  & $0.0$ & $948.6 \pm 12.1$ & $974.4 \pm 6.2$  & $925.4 \pm 17.2$ \\
  & $0.1$ & $871.4 \pm 19.8$ & $902.7 \pm 15.3$ & $838.1 \pm 26.4$ \\
  & $0.2$ & $742.6 \pm 34.5$ & $781.5 \pm 29.1$ & $695.3 \pm 41.8$ \\
\midrule
\multicolumn{5}{l}{\textit{Payload Mass} ($p$)} \\
\multirow{3}{*}{PME-PPO}
  & $0.0$ & $530 \pm 8.2$    & $602 \pm 14.9$   & $337 \pm 23.1$ \\
  & $0.1$ & $418 \pm 19.7$   & $455 \pm 24.3$   & $258 \pm 29.4$ \\
  & $0.2$ & $309 \pm 31.2$   & $298 \pm 34.1$   & $181 \pm 35.2$ \\
\midrule
\multirow{3}{*}{\ourMethod}
  & $0.0$ & $948.6 \pm 12.1$ & $974.4 \pm 6.2$  & $925.4 \pm 17.2$ \\
  & $0.1$ & $889.3 \pm 17.2$ & $896.1 \pm 16.7$ & $871.7 \pm 22.8$ \\
  & $0.2$ & $798.4 \pm 28.9$ & $769.3 \pm 31.4$ & $779.2 \pm 35.6$ \\
\bottomrule
\end{tabular}
\end{wraptable}

\textbf{Why not plan with the analytical model?} Since Analytical-$\mu$ is near-exact at nominal, one might use it for planning inside an MPC. The binding constraint, however, is correctness rather than speed. \autoref{tab:unmodeled_nmse} shows the analytical prior is \textit{wrong} under unmodeled dynamics, so no amount of planning recovers control once the robot departs from spec, and repairing it requires exactly the online SysID loop that our latent state amortizes away. A secondary practical point is that a per-morphology analytical model must be maintained for each robot in the control loop, whereas \ourMethod serves the whole cohort through $\mu$ injection into a single frozen forward model. The correctness axis is the one that cannot be engineered around.

\textbf{Bracketing both directions of spec-versus-reality divergence.} Taken together, \aref{app:mu_perturbation} and this section show that \ourMethod is robust along both axes of deviation from the idealized description, an inaccurate spec of a correct robot (corrupted $\mu$, nominal world) and a correct spec of a degraded robot (nominal $\mu$, perturbed world). A pure spec-sheet system, whether a model-free $\mu$-policy or a static analytical prior, is robust to neither, since its behavior is determined entirely by a $\mu$ that no longer matches reality. This is the empirical content of the claim that $(h_t, z_t)$ encode physical grounding beyond what $\mu$ alone provides, learned jointly without a differentiable physics engine in the loop.

\subsection{Implementation Details}
\label{app:unmodeled_protocol}

All experiments are run in the Hetero-Isaac environment, identical to training. Policies and WM weights are frozen.

\textbf{Torque attenuation.} The policy outputs joint position targets, which the PD controller converts to joint torques $\tau$. We scale the resulting torque, $\tau^{\text{eff}}=(1-\eta)\tau$, before it is applied to the articulation, leaving the robot description-declared effort limits (and hence $\mu_{\text{act}}$ in $\mu$) unchanged. $\eta$ is fixed per episode (resampled per draw), applied uniformly to all 12 joints.

\textbf{Payload.} We increase the base-link mass by a fraction of the robot's nominal total mass and update the link inertia consistently, placing the added mass at the trunk attachment point to produce a center-of-mass offset. The geometry and all other links are unchanged. $\mu$ is recomputed from the unchanged robot description and therefore remains nominal.

\textbf{Analytical-$\mu$ predictor.} Analytical-$\mu$ is a separate Hetero-Isaac instance configured with the robot's true nominal parameters, namely robot description-derived geometry and mass plus the friction and damping values used for the nominal robot (\aref{app:env_details}). It receives no information about the test-time perturbation. For each evaluation rollout we run \ourMethod's policy on the perturbed true robot, record the executed action (position-target) sequence and the ground-truth state trajectory, and initialize both predictors from the ground-truth state at $t=T=5$. Each predictor then rolls 45 steps open-loop fed only the recorded targets. Analytical-$\mu$ converts targets to torques via the same PD controller and integrates rigid-body dynamics. \ourMethod predicts via its $\mu$-conditioned transition model. No observation feedback is provided to either predictor during the rollout.

\textbf{NMSE.} We compute NMSE over the predicted physical proprioceptive channels, excluding the command and previous-action channels, which are inputs rather than predicted quantities. Errors are normalized per-channel by that channel's natural variance over the evaluation set, then averaged.